\RequirePackage{fix-cm}
\documentclass[smallextended]{svjour3}       % onecolumn (second format)
\smartqed  % flush right qed marks, e.g. at end of proof
\usepackage{graphicx}
\usepackage{subfig}
\usepackage{url}
\usepackage[autolanguage,np]{numprint}
\newtheorem{mydef}{Definition}
\usepackage{amsmath}
\usepackage{color}
\usepackage{multirow}
\usepackage{xcolor}
\definecolor{legend2-a}{HTML}{007FFF}
\definecolor{legend2-b}{HTML}{7FFF83}
\definecolor{legend2-c}{HTML}{FF8300}
\definecolor{legend2-d}{HTML}{830000}

\definecolor{legend1-a}{HTML}{002BFF}
\definecolor{legend1-b}{HTML}{00D7FF}
\definecolor{legend1-c}{HTML}{7FFF83}
\definecolor{legend1-d}{HTML}{FF2F00}

\newcommand{\mycaption}[2]{\multirow{#1}{*}[0.2em]{\fcolorbox{black}{#2}{\rule{0pt}{6pt}\rule{10pt}{0pt}}}}

\usepackage[utf8]{inputenc}
\usepackage[pagebackref=true,breaklinks=true,letterpaper=true,colorlinks,bookmarks=false]{hyperref}

\begin{document}

\title{3D Geometric salient patterns analysis on 3D meshes%\thanks{Grants or other notes
}


\author{
Alice Othmani \and 
       Fakhri Torkhani \and      
                         Jean-Marie Favreau 
                           }


\institute{A. Othmani \at
              Clermont Universit\'{e}, Universit\'{e} d'Auvergne, ISIT, BP 10448, F-63000 Clermont-Ferrand\\ 
              \email{Alice.Othmani@u-pec.fr}           %  \\
           \and
           F. Torkhani \at
              Clermont Universit\'{e}, Universit\'{e} d'Auvergne, ISIT, BP 10448, F-63000 Clermont-Ferrand\\
           \and
           J. M. Favreau \at
            Clermont Universit\'{e}, Universit\'{e} d'Auvergne, ISIT, BP 10448, F-63000 Clermont-Ferrand\\
           \email{j-marie.favreau@uca.fr}           
           }

\date{Received: date / Accepted: date}

\maketitle

\begin{abstract}
Pattern analysis is a wide domain that has wide applicability in many fields. In fact, texture analysis is one of those fields, since a texture is defined as a set of repetitive or quasi-repetitive patterns. Despite its importance in analyzing 3D meshes, geometric texture analysis is less studied by geometry processing community. This paper presents a new efficient approach for geometric texture analysis on 3D triangular meshes. The proposed method is a scale-aware approach that takes as input a 3D mesh and a user-scale. It provides as a result a similarity-based clustering of texels in meaningful classes. Experimental results of the proposed algorithm are presented for both real-world and synthetic meshes within various textures. Furthermore, the efficiency of the proposed approach was experimentally demonstrated under mesh simplification and noise addition on mesh surface. In this paper, we present a practical application for semantic annotation of 3D geometric salient texels.

\end{abstract}

\section{Introduction}
\label{sec:Introduction}
3D triangular meshes are widely used in computer graphics to present visual data to users. Such representations of 3D visual content can be found in several applications such as entertainment, cultural heritage, medical applications, or Computer Aided Design. In many cases, 3D mesh surfaces contain several geometric repetitive patterns in different scales. Such spatial details represent different regions with distinct characteristics on the mesh. Repetitive patterns on 3D meshes, called geometric texture, reflect the nature of surface shape aspect which is an important element to segment, recognize or annotate 3D mesh surface regions. It is therefore relevant to provide new algorithms that correctly segment and identify different textures with respect to a given scale. 

We note that several existing mesh processing tools attempt to quantify the local characteristics of mesh surfaces (e.g.\ quantify the smoothness or roughness of regions) or the global appearance of regions (e.g.\ planarity or sphericity of the surface). It could be possible to use such tools to separate and segment regions according to ad-hoc methods. Nevertheless, such algorithms are not designed to detect and extract repetitive structural patterns like geometric 3D texels. We emphasize the importance of the scale \cite{Uhl:2015} which is strongly linked to the 3D mesh density: within different scales, a geometric detail or noise will not be perceived the same way \cite{Lavoue:2009}. 

In this context, we present a new interactive method to segment and identify geometric textures on 3D mesh surfaces according to a user-defined scale. The proposed approach is efficient for both real-world and synthetic 3D textured meshes. The rest of the paper is organized as follows. In section~\ref{sec:contextandMotivations} we present related work and our motivations of this work. In section~\ref{sec:approach} we present the details of our algorithm to extract and classify 3D texels. In section~\ref{sec:ResultsandApplications} we present and evaluate our approach results and we introduce a practical application for semantic annotation of geometric textures. 

\section{Context and motivations}
\label{sec:contextandMotivations}
The main motivation of this work is to segment and extract regions having similar geometric details. In geometry texture analysis, these regions are usually called texels. 
There exists a wide range of possibilities to extract such geometric details from mesh surfaces, from the frenquency-based approaches separating high-frequencies from the base shape to approaches which segments similar regions on 3D meshes. 
An overview of the existing approaches is presented in the following section, and their limitations are discussed as an introduction to our motivations. We finally present as a realization of those different motivations a definition of geometric textures.
\subsection{Related work}

The prerequisite to analysing geometric textures consists on separating geometric details from the base mesh that represents the basic geometric shape \cite{Lai:2005}. To achieve this separation, frequency-based approaches are converting the geometry of a mesh into frequency space using low-pass filters \cite{Taubin:2000}, eigenfunctions of the Laplace-Beltrami operator \cite{Vallet:2008} or log-Laplacian spectrum \cite{Song:2014}. 

Another family of approaches defines geometric textures using height maps \cite{Lai:2005,Toledo:2008,Andersen:2009}. In these methods, the geometry texture superimposed on the object's base shape is defined as a vector of displacements \cite{Lai:2005} or in terms of rotation and displacement lengths from the normal vector \cite{Andersen:2009}. These geometric approaches are using a smoothing process and a vertex-to-vertex correspondence between the original textured surface and the smoothed version to produce the displacement or height map. One major limitation of methods presented above, with respect to our application, is the fact that they are not able to practically localize nor characterize the texture elements (texels).

We have also to consider a third family of approaches, strongly related to geometry textures. They encode regions of 3D mesh surfaces into a sparse set of local surface descriptors \cite{Gal:2006,Cheng:2007,Toledo:2008,Guy:2014,Othmani:2013}. In these works, coarse segmentations close to superpixels are presented, such as in \cite{Toledo:2008}, or more finely salient geometric features are extracted \cite{Gal:2006,Cheng:2007,Othmani:2013}. These features are defined by a set of descriptors, locally describing a non trivial region of the surface. These features can be identified by considering regions with a high curvature relatively to their surrounding and with a high variance of curvature values \cite{Gal:2006,Cheng:2007}. These purely geometric approaches can be enriched by the use of an optimized type-aware user-selection tool \cite{Guy:2014}.

\subsection{Motivations}

We have presented, in the previous section, an overview of existing approaches addressing geometric textures analysis. According to our knowledge in this domain, existing approaches are not able to finely and precisely demarcate the boundaries of the geometric details, they only give a coarse representation of mesh features preventing a precise characterization of the fine texture details. Moreover, we have noticed that none of these methods are considering the question of the scale, neither the question of the region orientation (i.e. distinguish between regions on the top and regions on the bottom of the base shape).

All those limitations are the basic motivation leading to conducting such a work and answering those different points leading to a more precise analysis of geometric textures that can target different domains, such as visualization~\cite{Toledo:2008}, rendering~\cite{Toledo:2008}, smoothing~\cite{Toledo:2008}, detail-replicating and shape stretching~\cite{Alhashim:2012}, texture mapping and synthesis~\cite{Zhou:2006}. One of the open applications to geometric texture analysis is the semantic annotation in order to reduce the semantic gap between the algorithm of image processing and the application. This application stands for the second motivation of this work.

The present article introduces an original approach of salient textures analysis on 3D triangular meshes, with a strong focus on the following characteristics:
a scale-aware and orientation-aware segmentation of fine geometric details, a clustering by geometric similarity of the geometric details to form the geometric textures, a texture characterization valuable for the annotation, and open-source diffusion of our implementation to the scientific community.

\subsection{Definitions}
\label{sect:definitions}

In the following, we consider 3D objects described by oriented closed meshes $\mathcal{M}=(V,E,F)$ defined by a set of vertices $V$ (defined by their 3D coordinates), a set of edges $E=\{ (i, j) \in V^2\}$ and a set of triangular facets $F=\{ \{e_i\}_i \subset E\}$. 

\begin{figure}
    \centering
    \includegraphics[width=\columnwidth]{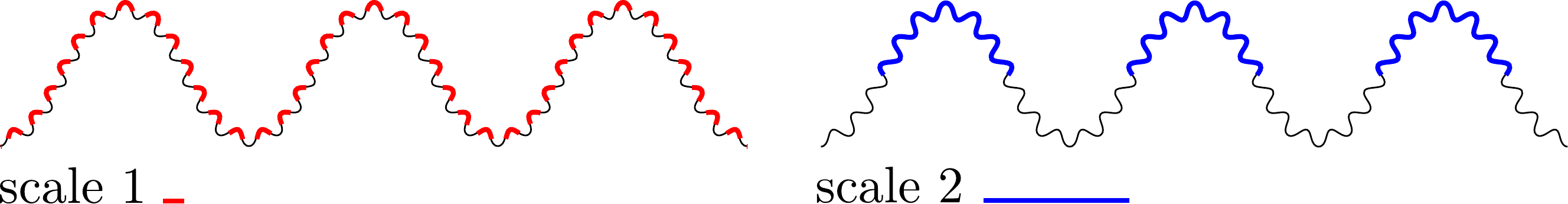}    
    \caption{Illustration of 3D texels for two different scales.}
    \label{fig:scale}
\end{figure}

To satisfy the motivations described in the previous section, we consider a \textbf{3D texel} on a mesh $\mathcal{M}$ as geometrically 3D salient region at a given scale.
Figure~\ref{fig:scale} illustrates the scale dependency of 3d texels. 
Thus a \textbf{geometric texture} on $\mathcal{M}$ is a set of 3D texels having similar geometric characteristics.

\section{3D geometric textures analysis}
\label{sec:approach}

\begin{figure}
    \centering
    \includegraphics[width=\textwidth]{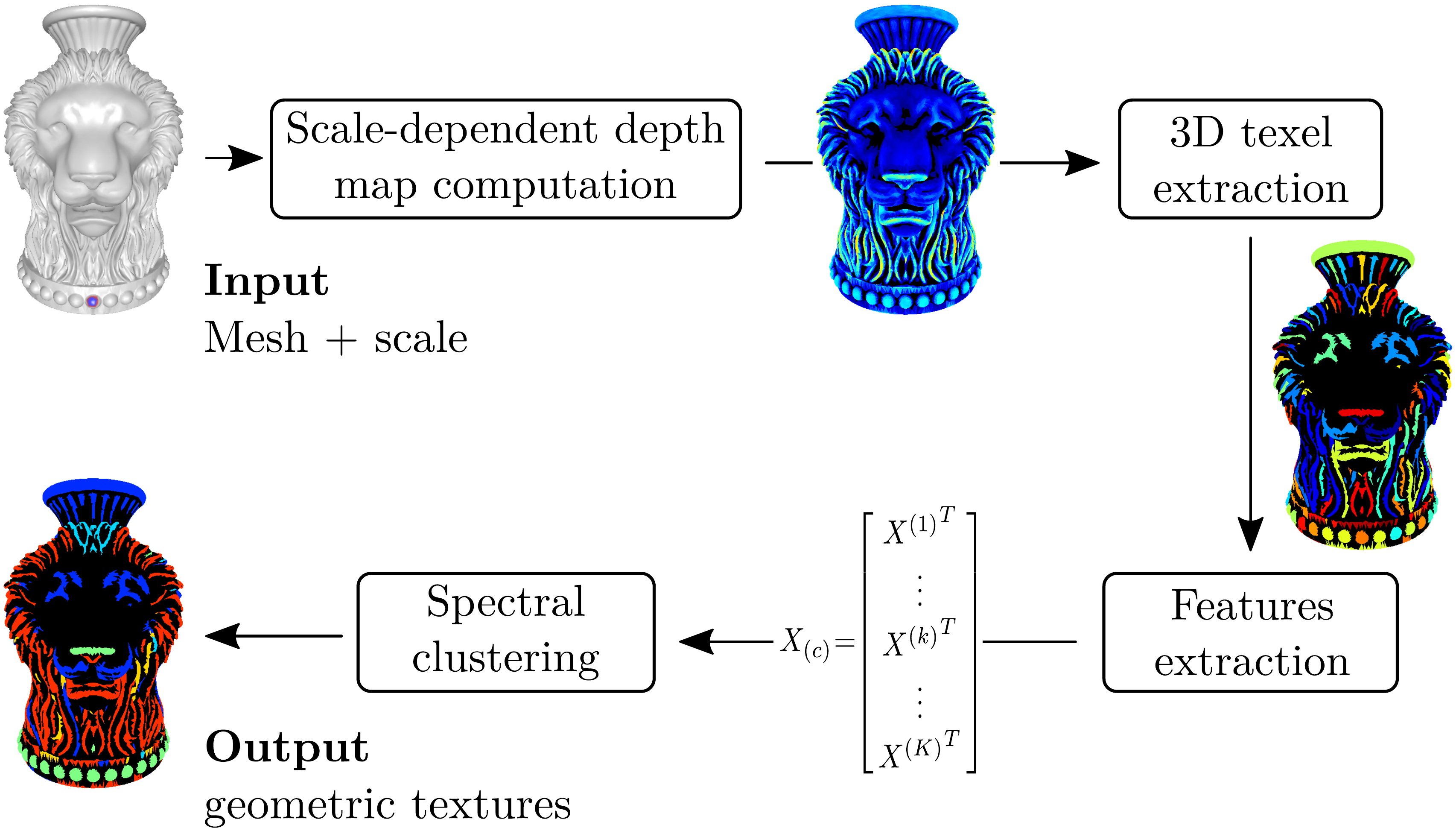}    
    \caption{Flow diagram of the approach.}
    \label{fig_1}
\end{figure}

The approach we present in this work aims to identify texels as regions significantly distinct from the global shape by computing first a scale-dependent depth map then a texels segmentation.
Each of these elements of geometry are then gathered together into geometric textures using geometric features to characterize their appearance.
\figurename~\ref{fig_1} illustrate the 4-step approach detailed in the following sections.

\subsection{Scale-dependent depth map computation}
\label{sect:depth}

\begin{figure}[]
        \centering
        \subfloat[A 2D shape describing an object with one ``positive detail'' (left) and one ``negative detail'' (right); the blue part corresponding to the inside of the object.]{\label{fig:smooth-shape} \includegraphics[width=.8\columnwidth]{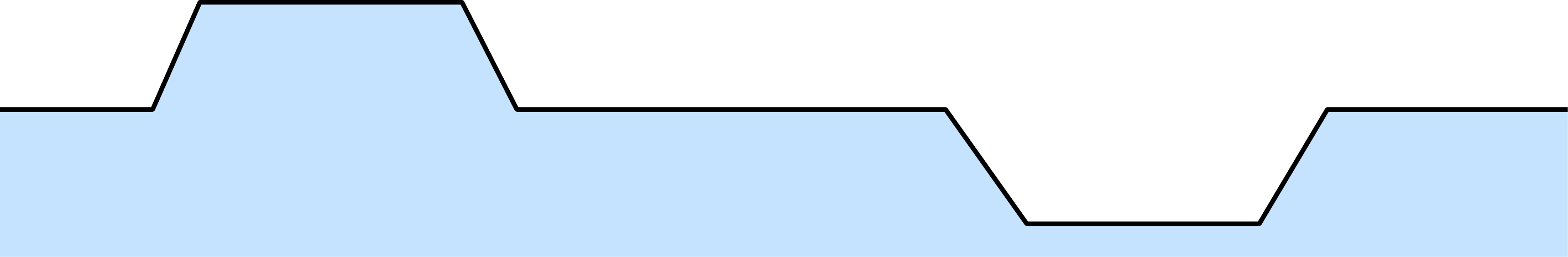}}              
                
        \hspace{5pt}
        \subfloat[Symmetric filtering. The red curve illustrates a classical high frequency filtering result. The bold black lines are materializing the significantly different regions, with respect to this filtering.]{\label{fig:smooth-symmetric}\includegraphics[width=.8\columnwidth]{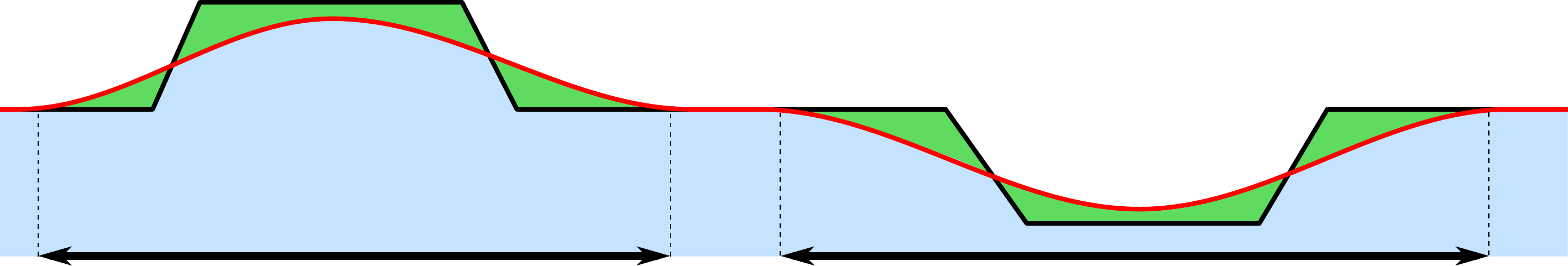}}
                
        \hspace{5pt}
        \subfloat[Oriented smoothing, preserving only positive details.]{\label{fig:smooth-positive}\includegraphics[width=.8\columnwidth]{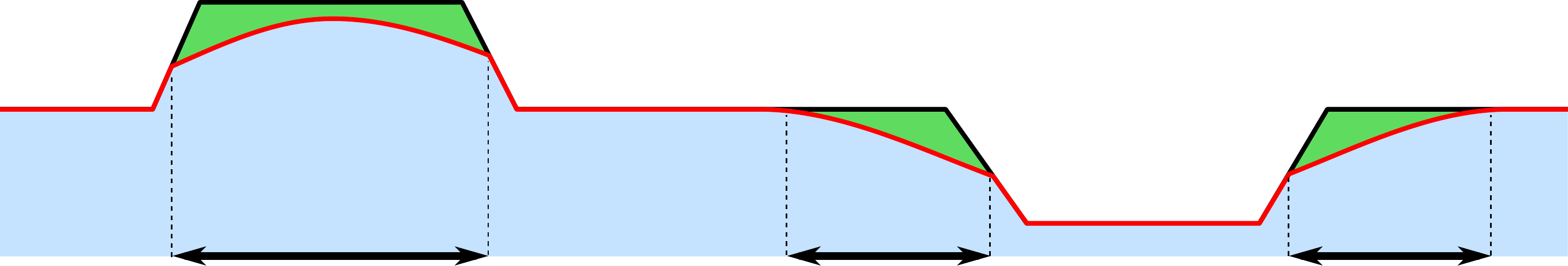}}
 
 \hspace{5pt}
        \subfloat[Oriented smoothing, preserving only negative details.]{\label{fig:smooth-negative}\includegraphics[width=.8\columnwidth]{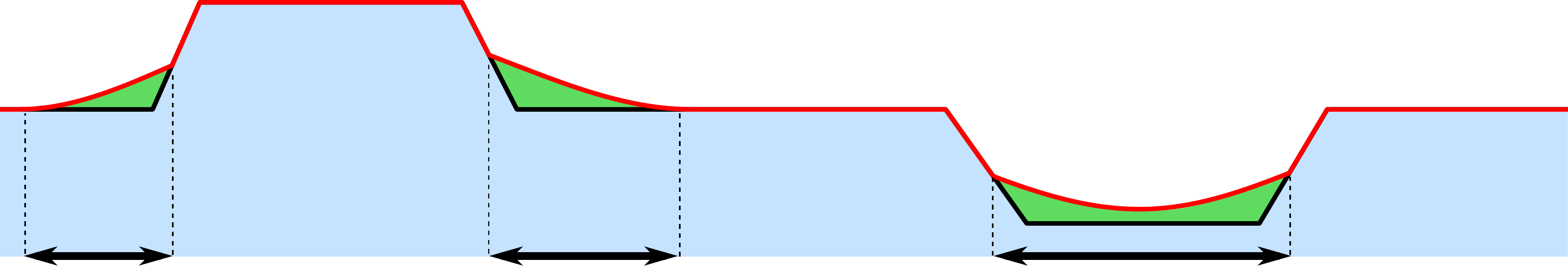}}
        
     \caption{Illustration of symmetric and non-symmetric smoothing on 2D profiles
         with small geometric details.}
    \label{Fig:smoothing}
\end{figure}

Almost all the approaches we described in section~\ref{sec:contextandMotivations} are identifying geometric textures as high frequency components of the shape. An illustration of spectral filtering result is presented in Figure~\ref{fig:smooth-symmetric}. 
% The significant regions identified using  this approach are filled out with a main consequence:
% loosing the accuracy in the region delimitation
% }{pas cool comme phrase, on remplace?: Different regions of the shape can be extracted imprecisely with this approach}. 
The delimitation of the significant extracted regions is imprecisely located.
In fact, considering that the shape is composed of a background (the default shape) and a foreground (the fine salient details), a symmetric smoothing will select not only the foreground details, but also the neighborhood (Figure~\ref{fig:smooth-symmetric}).

On the contrary, working with a non symmetric operation as suggested in Figure~\ref{fig:smooth-positive} and \ref{fig:smooth-negative}, we can extract with a good accuracy the salient details only in a given direction: 
\emph{positive details} are regions on the top of the low frequency shape, while \emph{negative details} are regions on the bottom of it. This observation is one of the key element of our work. 

In the following, we describe a non-symmetric and scale-dependent depth map computation. This process associates to each vertex of the original mesh a scalar value that declares if this vertex is part of a positive detail of the shape at the given scale.

\subsubsection{Scale-dependent Laplacian smoothing}

The depth map computation is based on the computation of a smoothed oriented version $\mathcal{M}_s^+$ of an original mesh $\mathcal{M}_o$. $\mathcal{M}_s^+$ can be considered as the global shape of the object, and the deviation between $\mathcal{M}_s^+$ and $\mathcal{M}_o$ as the positive details.

In order to provide a scale dependent approach, we perform a Laplacian smoothing on the original mesh, computing for each vertex $v_o$ of $\mathcal{M}_o$ the corresponding vertex $v_s$ as the barycenter of its neighborhood at a given scale $r$:
\begin{equation}
 v_s= \frac{1}{\left|N_r(v_o)\right|}\sum_{v_i \in N_r(v_o)} v_i.
\end{equation}
In this equation, $N_r(v_o)$ is the neighborhood of $v_o$ defined as the set of all the vertices in the geodesic disc around $v_o \in \mathcal{M}_o$ contained in the sphere of radius $r$ centered in $v_o$.

The result of this operation corresponds to a symmetric smoothing (Figure~\ref{fig:smooth-symmetric}). In order to preserve only positive details, we consider the normal $n(v_o)$ to compute the smoothed oriented mesh $\mathcal{M}_s^+=(\{v_s^+\}, E, F)$ from the original mesh $\mathcal{M}_o=(\{v_o\}, E, F)$:

\begin{equation}\label{eq:nonsymdev}
 v_s^+ = \left\{  \begin{array}{ll}
                    v_s &\mbox{if } (v_o - v_s) \cdot n(v_o) > 0\\
                    v_o & \textrm{otherwise}.
                   \end{array}
\right.
\end{equation}
In this article, we only focus our study in $\mathcal{M}_s^+$, but a similar definition $\mathcal{M}_s^-$ can be written for negative details. The orientation depends here only on the direction of the normal.
\subsubsection{Oriented depth map computation}

Once a smoothed oriented version of the mesh has been computed, we generate the depth map as the geometric deviation between the original mesh $\mathcal{M}_o$ and a smoothed version $\mathcal{M}_s^+$.
For this purpose, we compute for each vertex $v_o$ of $\mathcal{M}_o$  the Euclidean distance $\|v_o - \tilde{v}_o\|$ between $v_o$ and $\tilde{v}_o$, where $\tilde{v}_o$ is the nearest neighbor of $v_o$ in the smoothed mesh $M_s^+$. $\tilde{v}_o$ is determined using the efficient Aligned Axis Bounding Box tree structure \cite{Alliez:2009}. In the following we write $D_r^+(v)$ the oriented depth at scale $r$ of vertex $v\in \mathcal{M}_o$.

\begin{figure}
 \hspace*{\stretch{1}}
 \raisebox{-0.5\height}{\includegraphics[width=0.26\linewidth]{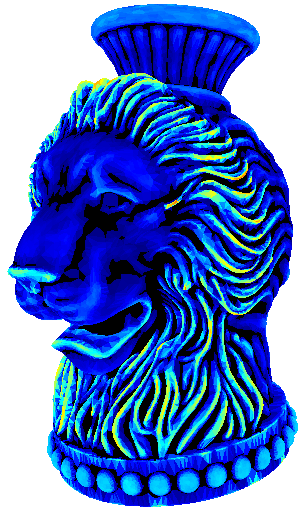}}
 \hspace*{\stretch{1}}
 \raisebox{-0.5\height}{\includegraphics[width=0.3\linewidth]{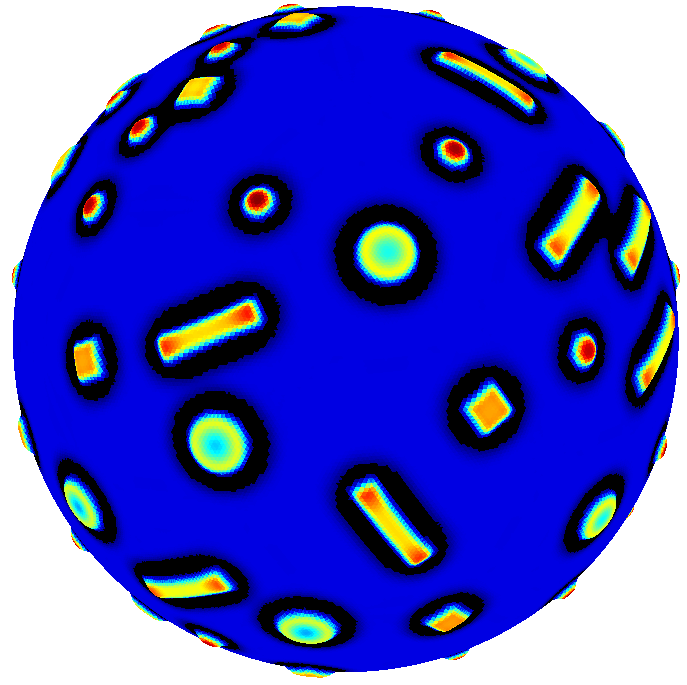}}
 \hspace*{\stretch{1}}
 \caption{Oriented depths computed on two meshes. Warmer colors represent larger values.}
 \label{fig:depthmap}
\end{figure}

By extension, we can compute a depth $D_r^+(f)$ for each facet $f$ by computing the mean of its vertices' depth: 
\begin{equation}
D_r^+(f)= \frac{1}{\{v_i\}_f} \sum_{\{v_i\}_f} D_r^+(v_i). 
\end{equation}

Figure~\ref{fig:depthmap} shows two depth map computed on 3D meshes.

\subsection{3D texels extraction}

In section \ref{sect:definitions} we introduced 3D texels as geometrically 3D salient region at a given scale. 
The oriented depth $D_r^+(\cdot)$ is an effective way to quantify for each vertex of the mesh its degree of saliency.  Thus a direct way to extract 3D texels consists on producing a segmentation of the mesh based on this oriented depth.

We identified that a global thresholding of the oriented depth map is not relevant to achieve this segmentation, in particular because the complex geometry of the shapes can produce 
low frequency variations on the depth map (see Figure~\ref{fig:depthmap}). To overcome this fact, we introduce in this section an adaptative thresholding dependent to the local depth distribution.

\subsubsection{Geometric details location}
The first step of the texel extraction is to find the location of the geometric details. Using the depth map as a scalar function on the mesh, we identify local maxima as \emph{seeds} of the 3D texels.

Local maxima of the scalar function are first computed considering that a vertex $v$ is a maxima if $D_r^+(v) > D_r^+(v')$ for all $v'$ contained in a geodesic disc around $v$ and in the sphere of radius $r$ centered in $v$.

The set of all possible seeds is finally filtered to preserve only elements being part of the salient regions of the global shape. In pratical, we use Otsu's method \cite{Otsu:1979} on the complete depth map to define a global threshold separating the salient regions from the others, and only select as final seeds the salient ones.

\subsubsection{Geometric details segmentation}

% Starting from the set $S$ of all final seeds, we generate the final texels by computing first one local region per seed, then by merging the connected local regions as a single texel. Since a texel is a subpart of the mesh, we choose to describe a texel as a set of facets, a facet being the elementary piece of surface. 

The local region $R(v_s)$ associated to a seed $v_s$ is a salient region at the given scale $r$. A near-neighborhood of radius $2r$ should then contain both salient and non salient pieces of surface.
We thus examine near-neighboring facets of $v_s$ inside a geodesic disc in the sphere of radius $2r$ centered in $v_s$. We separate these neighboring facets into two classes so that their combined depth spread is minimal and their inter-class depth variance is maximal. We use here Otsu's method \cite{Otsu:1979} to estimate the optimum threshold $t(v_s)$ for each vertex $v_s$. The local region $R(v_s)$ associated to $v_s$ is then generated by computing the connected component of $v_s$ on $\mathcal{M}_o$ such that $D_r^+(v) \geq t(v_s)$ for each facet $v \in R(v_s)$. 

Once this process has been done for each seed, some pair of regions can be adjacent or even share a common triangle. In that case, we choose to merge or not these regions using an evaluation of their similarity in term of depth. This evaluation is done by estimating the overlapping between the two depth distributions. More precisely, we compute for each region $R(v_{s_i})$
the mean $\mu_i$ and standard deviation $\sigma_i$ of its depth, and compare the overlapping width of the two intervals $\left]\mu_0 - \sigma_0; \mu_0 + \sigma_0\right[$
and $\left]\mu_1 - \sigma_1; \mu_1 + \sigma_1\right[$ with the width of the intervals.

At the end of this merging step, we obtain a series of regions $\mathcal{R}=\left\{R_0, ..., R_m\right\}$ called 3D texels, corresponding to the geometric details of the input mesh.
\subsection{Features extraction}
\label{sec:features}

Once the texels have been extracted, the next step consists on associating features that are describing the shape to each of these small regions. These features will be used as the input of an unsupervised classifier to group regions with similar aspect.

We can classify these features in three categories: features that describe the shape of the contour of the regions, features that describe the global appearance of the shape, and features that describe the local region aspect.

\subsubsection{Contour features}
The first set of features we introduce corresponds to descriptors associated to the contour $C=(v_0, ..., v_n)$ of a region. Notice that to compute these features we first applied a smoothing on the contour $C$ of each region to avoid noisy contours due to the structure of the triangular mesh. The considered contour $C'$ is obtained by adjusting the location of each contour vertex as following:
\begin{equation}
 v'_i = \frac{1}{4}(v_{i-1} + 2 v_i + v_{i+1}),
\end{equation}
where $v_{i-1}$ and $v_{i+1}$ are $v_i$'s neighbors along the contour.

\begin{figure}
 \centering
 \raisebox{-0.5\height}{\includegraphics[width=.35\columnwidth]{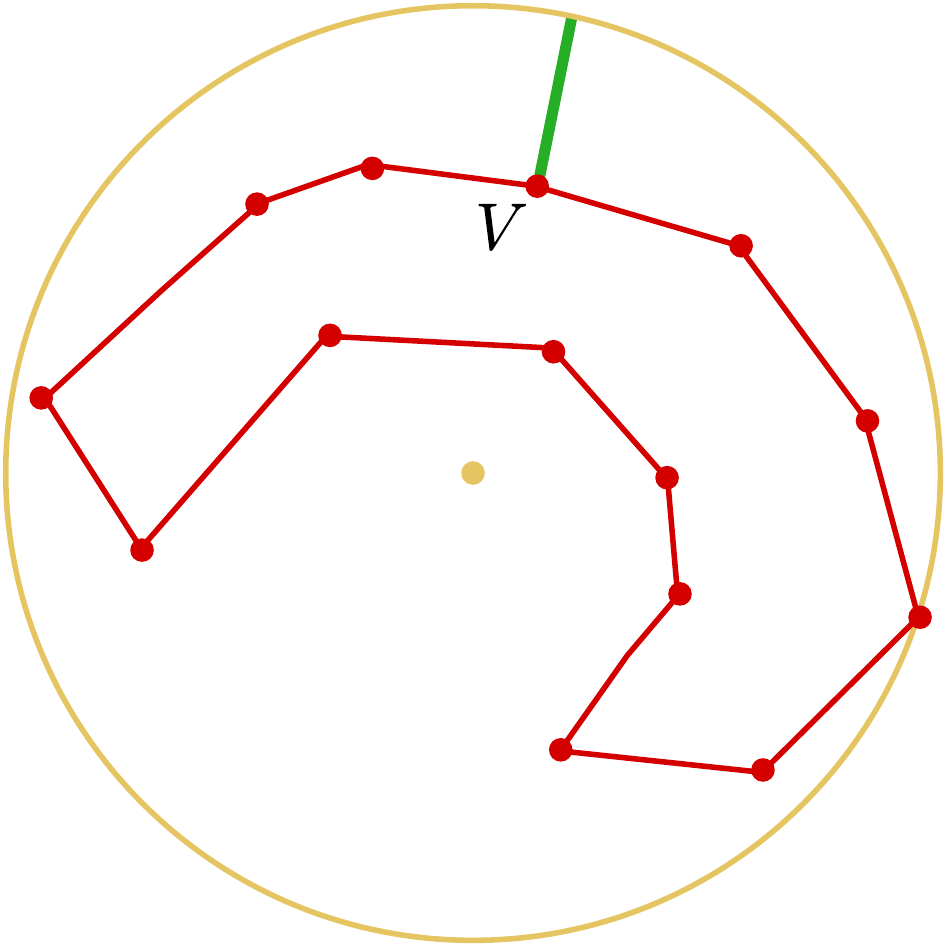}}
 \hspace{\stretch{1}}
 \raisebox{-0.5\height}{\includegraphics[width=.55\columnwidth]{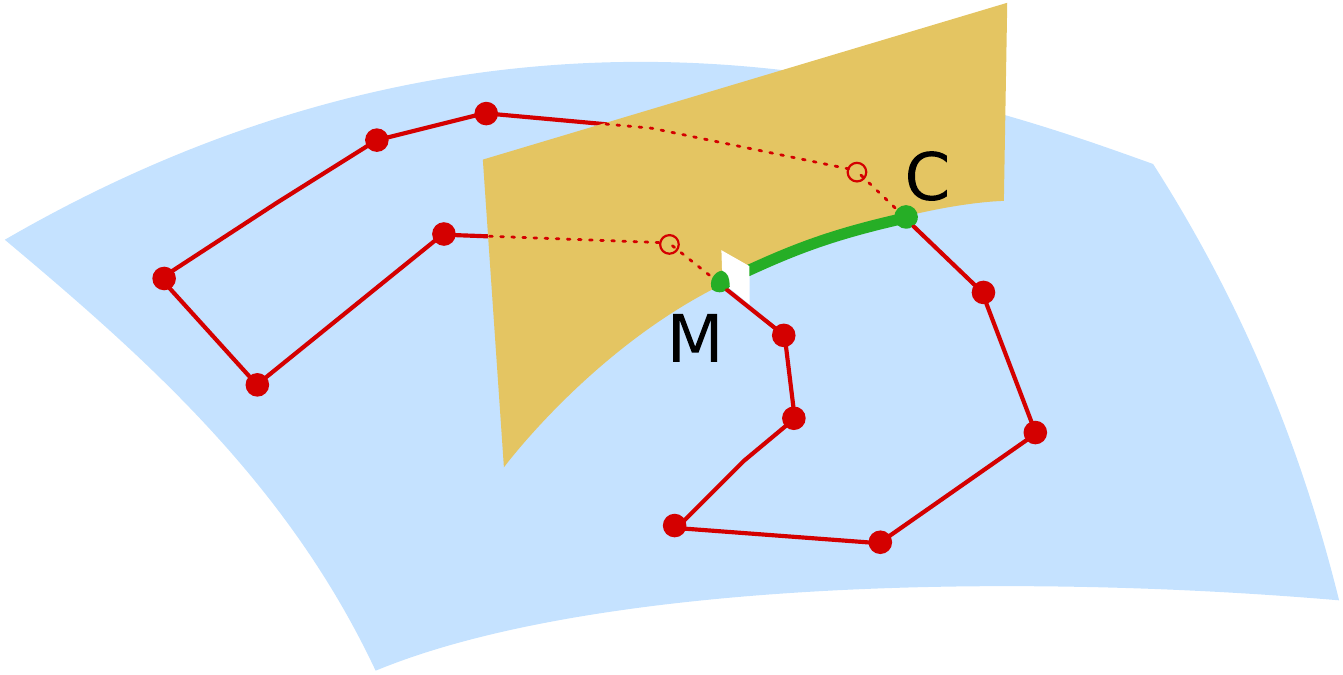}}
 \caption{Contour features inspired from classical shape descriptors. Original contour in red.
 Left: distance to a bounding sphere (yellow). Right: local diameter of the shape at point $M$ using a bisecting plane (yellow).}
 \label{Fig:ContourFeatures}
\end{figure}

% Perimeter (ok)
The first basic feature is the length of the perimeter. We also introduce two different features inspired from classical shape descriptors. 
\paragraph{Contour sphericity.}
% Sphericity (ok)
In order to estimate how rough the contour is, we introduce a first feature to estimate how far the contour is from the bounding sphere $C$ centered in the barycenter of all vertices in the region. Our feature consists on computing for each vertex of the contour its distance to this sphere (Figure~\ref{Fig:ContourFeatures} left), and by considering the standard deviation of this distance.  

\paragraph{Local diameter.}
% LocalDiameter_Std_Deviation LocalDiameter_Mean (ok)
The last features we introduced are inspired by the Shape Diameter Function~\cite{Shapira:2008} in a very basic way. For each edge of the contour, we estimate a local diameter of the region by computing the intersection between the tangent plane and the set of all the other contour edges (Figure~\ref{Fig:ContourFeatures} right). 
We compute for each contour the mean and standard deviation of this local diameter. Notice that we normalized these values using the maximum local diameter over all the contour edges of all the contours.

\subsubsection{Appearance features}
 
Beyond the contour descriptors, the global appearance of the regions as 3D surfaces has also to be described by features. We introduce here some classical features related to 3D shape description, as a property of these regions. The first appearance features we introduce are the area and the bounding sphere radius, to handle the size of the region. These features has been normalized between 0 and 1 using maximum and minimum values over all the regions. Another appearance feature we introduce is the squared perimeter to area ratio. In case of flat shape, it corresponds to a classical shape descriptor circularity ratio. In case of a curved surface, it also contains a characterization of this local curvature~\cite{Mortara:2004}.

Finally, we introduce features computed from the Principal Component Analysis (PCA) of the 3D coordinates, to encode the global aspect of the shape within the three principal directions of the region, with the three eigenvalues of the PCA normalized by the overall sum.

\subsubsection{Local aspect features}

The aspect of a 3D shape is usually measured using local geometric features, using technics such as distributions as a signature of the object~\cite{Osada:2002}. The description of these distributions has to be adapted to our application. In this context, the texel regions can contain a small number of vertices, and we have to be able to compare efficiently the distributions to cluster texels with respect to their aspect features. Moreover, the main application we present (see section~\ref{sec:annotation}) consists on a semantic description based on these features. To achieve these goals and satisfy the constraints, we choose to describe the distribution of local features using only the mean and standard deviation of these distributions.

The first local aspect descriptor we introduced is based on the depth of the vertices. This depth is encoding the local gap between a region and the global smooth shape. Once we normalized this depth between 0 and 1, the mean of a region describe if it is high or low, since the standard deviation encodes the internal variation of depth. Local curvature operators has been identified \cite{Lee:2005} as good criteria to describe a shape. We thus integrate in the local aspect descriptors the mean and standard deviations from the Gaussian curvature, shape index and curvature index~\cite{Koenderink:1992}. The intuitive understanding of these features is presented in section~\ref{sec:annotation}.

\subsection{Self-tuning spectral clustering}
\label{sec:spectral-clustering}

Starting from a 3D mesh, we first extracted 3D texels $\mathcal{R}=\left\{R_0, ..., R_m\right\}$ and we associated to each texel $R_i \in \mathcal{R}$ a series of values $\left\{f_{i,k}\right\}_k$ corresponding to geometric features (section~\ref{sec:features}). The last step of the pipeline consists on clustering similar texels into geometric textures (see definition on section~\ref{sect:definitions}). The clustering problem is related to finding a partition $\left\{ \mathcal{T}_i \subset \mathcal{R}\right\}_i$ of $\mathcal{R}$ such that regions in a \emph{class} or \emph{geometric texture} $\mathcal{T}_i$ are significantly similar while being significantly non similar with the regions of the other geometric textures. 

We decided to use in this work spectral clustering, which has many fundamental advantages compared to ``traditional algorithms'' such as k-means: it is simple to implement and it can be solved efficiently by standard linear algebra methods. 
In practice, spectral clustering solves a spectral graph partitioning problem. First a similarity matrix $(A_{i,j})_{i, j}$ is computed. Each $A_{i,j}$ derivates from the Euclidean distance between features $\left\{f_{i,k}\right\}_k$ and $\left\{f_{j,k}\right\}_k$ and encodes the similarity  between the two regions $R_i$ and $R_j$.
This similarity matrix is then used to perform dimensionality reduction before clustering in fewer dimensions.

Classical spectral clustering methods require as user-defined input the targeted class number, and have some constraints related to the uniformity of graph density. L. Zelnik-Manor and P. Perona proposed in 2004 a self-tuning spectral clustering \cite{LihiZelnikManorPPerona:2004} that solves these different issues. In this approach, the affinity between each pair of points is defined using a local scale and the structure of the eigenvectors is exploited to infer automatically the number of clusters, as a result of a better clustering especially when the data includes multiple scales and when the clusters are placed within a cluttered background.

In our application, the number of geometric textures to study is not limited, with no guarantee on the distribution uniformity of regions in the feature space. Moreover, we are looking for an unsupervised approach where the user is not required to give final geometric textures number. Applying the self-tuning spectral clustering introduced by L. Zelnik-Manor and P. Perona satisfies all these requirements, and provides us a straightforward method to produce geometric textures from the original set of 3D texels.
\section{Results and applications}
\label{sec:ResultsandApplications}

We have implemented mesh processing algorithms of the presented pipeline using a \texttt{C++} framework based on CGAL\footnote{CGAL library: \url{http://www.cgal.org/}}, and we used an existing implementation\footnote{Self-tuning spectral clustering implementation:\\ {\url{http://www.vision.caltech.edu/lihi/Demos/SelfTuningClustering.html}}} of the self-tuning spectral clustering. All the implementation is published under open source license\footnote{Source code of our work and instructions to use it are available on \url{https://github.com/AliceOTHMANI/3D-Geometric-Texture-Segmentation}}.

We applied our segmentation and classification pipeline to several meshes to illustrate the relevance of this approach. In this section, we present first a series of experiments to illustrate the robustness and accuracy of the method, both on the segmentation and classification part. In a second subsection, we present segmentation and classification results on meshes obtained from the AIM@Shape project. In the last part of this section, we introduce a first application of the current pipeline, as a process to produce semantic annotation from 3D mesh.

\subsection{Evaluation}
\label{sec:evaluation}
\begin{figure}
 \includegraphics[height=.33\columnwidth]{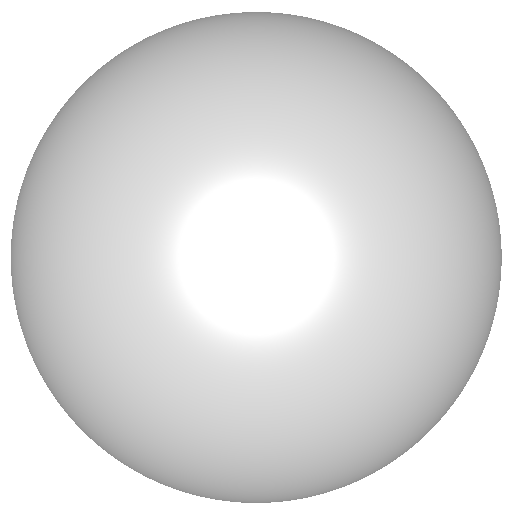}
 \hspace{\stretch{1}}
 \includegraphics[height=.33\columnwidth]{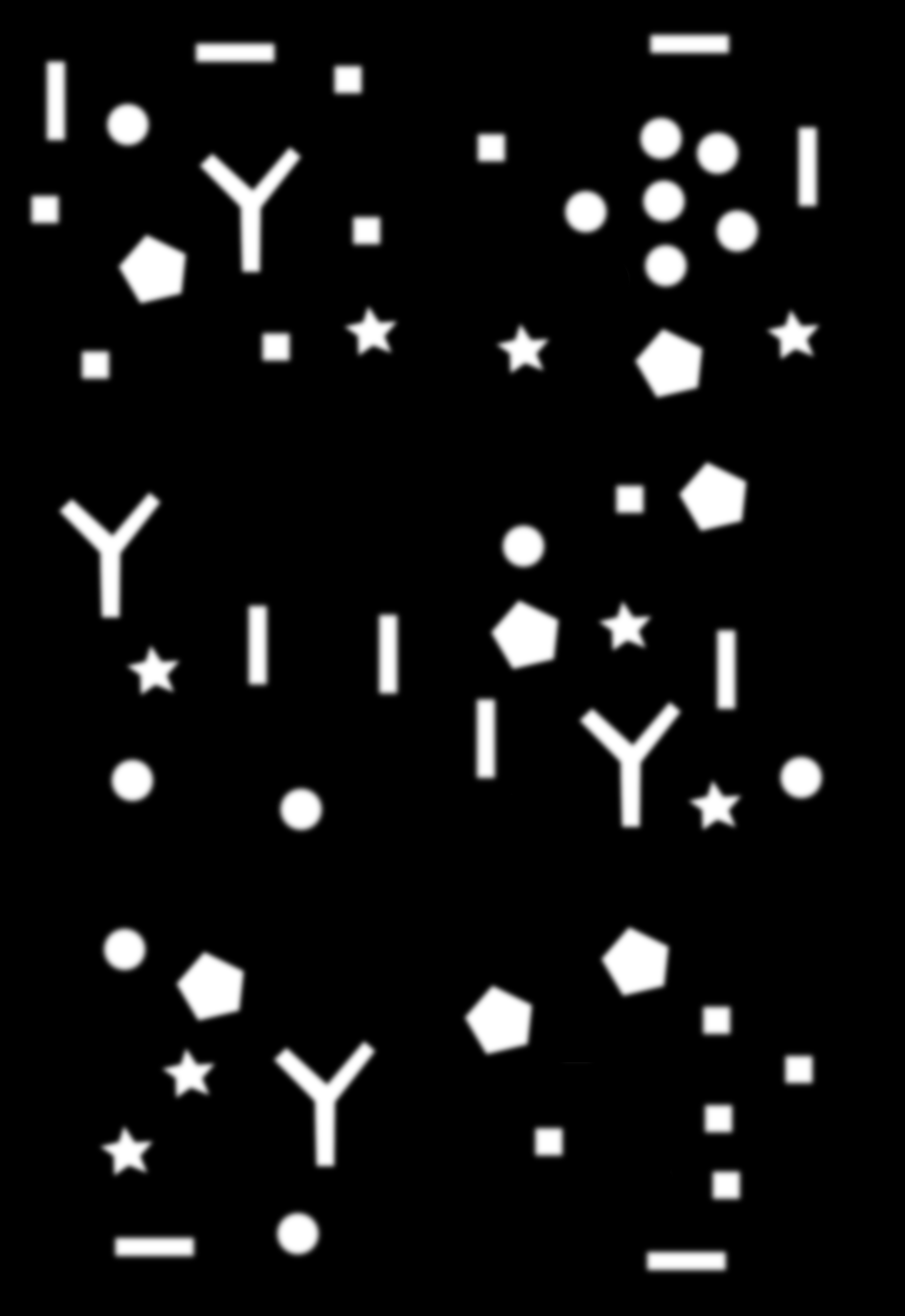}
 \hspace{\stretch{1}}
 \includegraphics[height=.33\columnwidth]{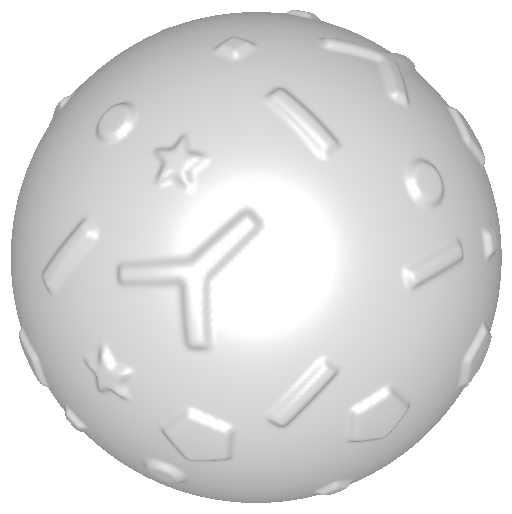}
 \caption{A sphere composed of \np{327680} faces, a greyscale image, and the result considering the image as a displacement map on the initial sphere.}
 \label{fig:disp-map}
\end{figure}

\begin{figure}
   % \centering
    \subfloat[Synthetic mesh 1: original mesh and segmentation result. Blue: big discs; green: small discs; red: squares; orange: rectangles.]{\label{fig:sphere-texture1} \includegraphics[width=.4\linewidth]{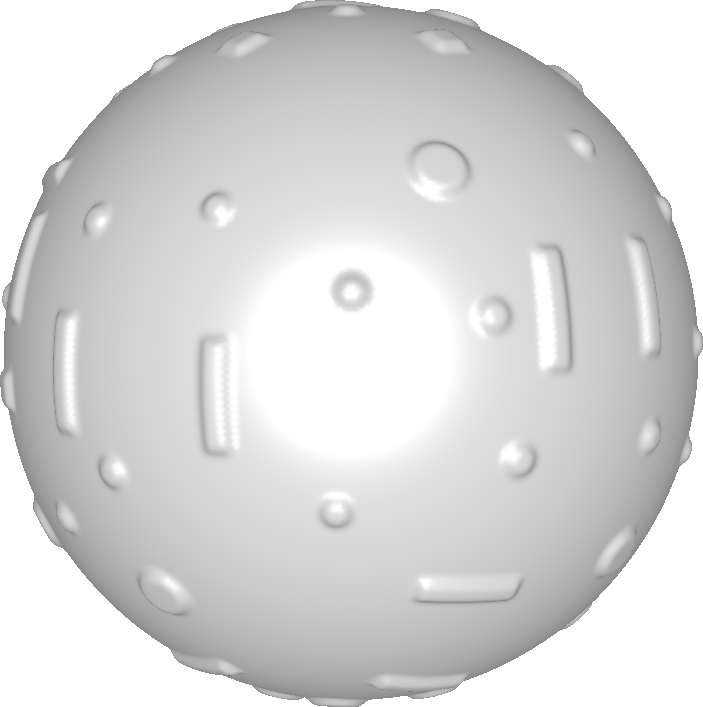} \hspace*{\stretch{1}}
\includegraphics[width=.4\linewidth]{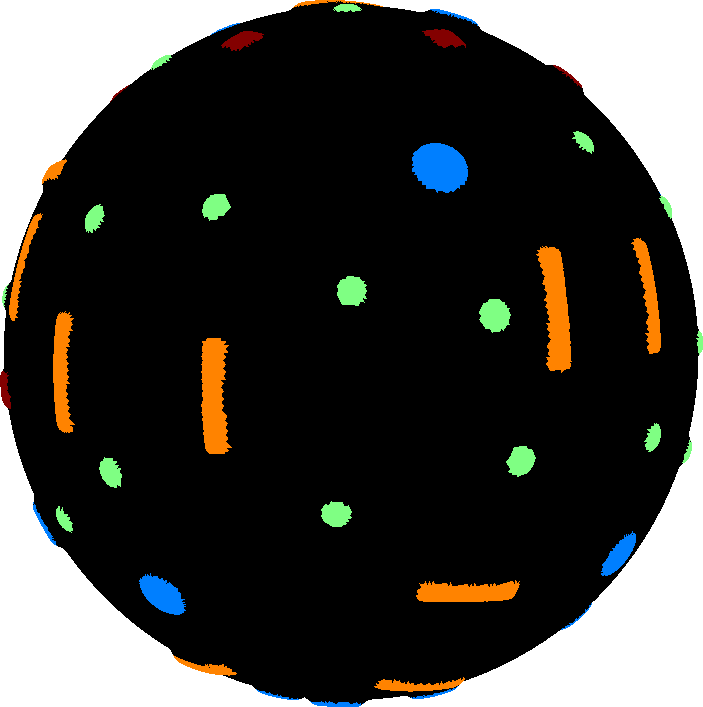}}

    \subfloat[Synthetic mesh 2: original mesh and segmentation result. Red: ``Y'' shape; green: rectangles.]{\label{fig:sphere-texture2} \includegraphics[width=.4\linewidth]{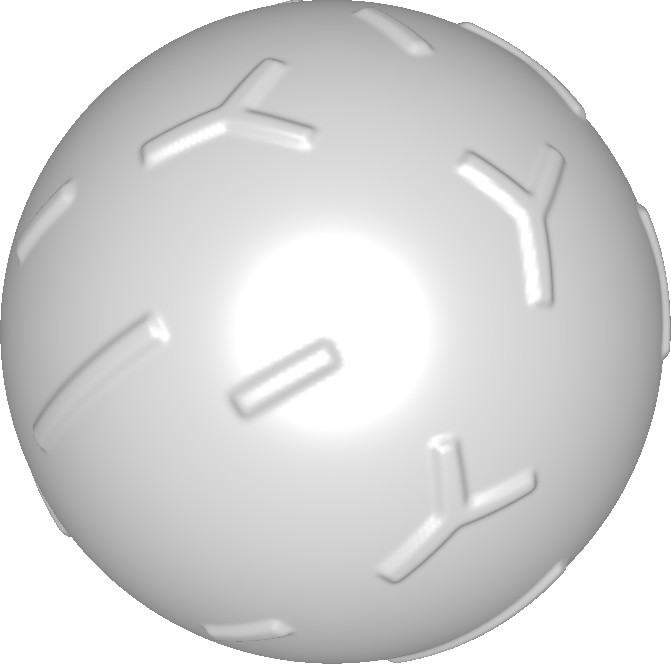} \hspace*{\stretch{1}}
\includegraphics[width=.4\linewidth]{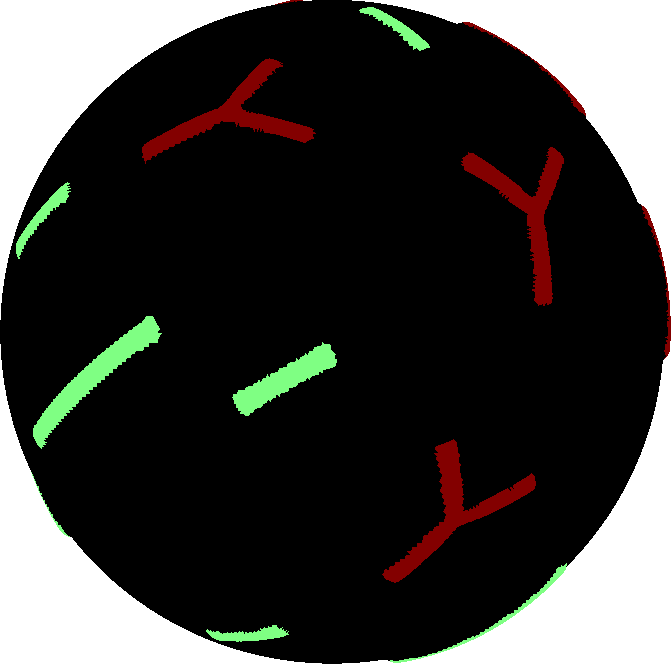}}

\subfloat[Synthetic mesh 3: original mesh and segmentation result.  Red: pentagons; green: stars.]{\label{fig:sphere-texture3} \includegraphics[width=.4\linewidth]{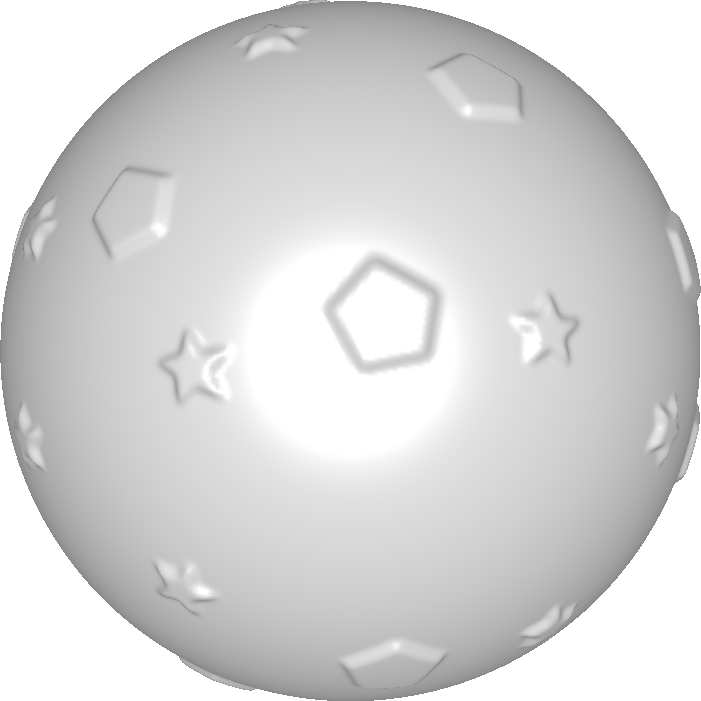} \hspace*{\stretch{1}}
\includegraphics[width=.4\linewidth]{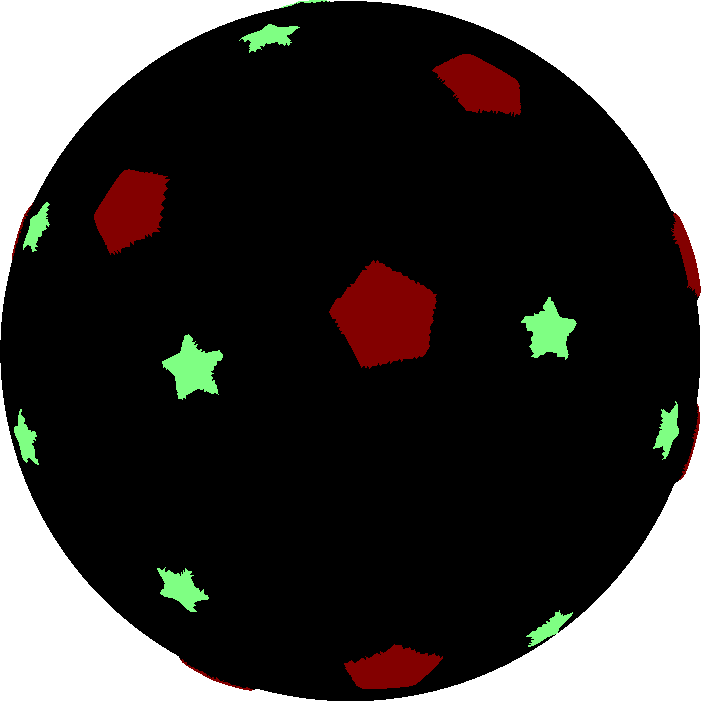}}

    \caption{Results of the classification on synthetic meshes. One color per class on each mesh.}
    \label{fig:synthetic-segmentation-intro}
\end{figure}

The evaluation of the method has been done using synthetic meshes generated from an initial icosphere (\np{327680} triangles) and a displacement map with geometrical details described by white pixels ($255$) on a black background ($0$), with all possible intermediate elevation. Figure~\ref{fig:disp-map} shows an example of displacement map applied on the initial sphere. Figure~\ref{fig:synthetic-segmentation-intro} shows the result of the segmentation and classification on other synthetic meshes. We applied here a maximum displacement corresponding to 2\% of the sphere radius.

The UV mapping that we applied to perform the displacement of the mesh has been used to transfer the grayscale value of the displacement map to the vertices of the mesh. This direct mapping gives us a way to produce a ground truth segmentation on these synthetic meshes: we associate to each facet $t$ the mean $g(t)$ of the grayscale values of its vertices, then we use a threshold $t=128$ to label each facet as ``ground truth foreground'' ($g(t) > t$) or ``ground truth background'' ($g(t) \leq t$).

\subsubsection{Hausdorff distance for segmentation evaluation}
\label{sec:hausdorff}
In the two next sections we present the robustness of the method by introducing noise and by reducing the resolution of the mesh. 
To evaluate the quality of the background/forgeround segmentation, we introduce here a surfacic Hausdorff distance, as a declination of the classical definition to surfacic regions.

\begin{mydef}
Let $\mathcal{M}=(V, E, F)$ be a mesh, and $S_0, S_1 \subset F$ two background/forgeround  segmentations of $\mathcal{M}$. The Hausdorff distance between $S_0$ and $S_1$ is defined as:

\begin{multline}
d_H(S_0, S_1)=\frac{1}{B_\mathcal{M}}\max \left\{ \max_{f_0 \in S_0} \min_{f_1 \in S_1} d(f_0, f_1), \right.
\left.\max_{f_1 \in S_1} \min_{f_0 \in S_0} d(f_1, f_0) \right\},
\end{multline}

where $d(f_0, f_1)$ is the Euclidean distance between $f_0$ and $f_1$ barycenters and $B_\mathcal{M}$ the maximum bounding box size.
\end{mydef}
The normalization by the maximum bounding box gives an Hausdorff distance proportional to the size of the object, to facilitate its interpretation.

\subsubsection{Influence of the noise}

The mesh we presented on Figure~\ref{fig:sphere-texture1} contains \np{327680} triangles, and we drawn on it 18 rectangles, 16 squares, 17 big discs and 37 small discs, with a typical diameter corresponding to $5\%$ of the maximum bounding box. We applied on this mesh a uniform noise with several maximum intensities, expressed in percentage of the sphere radius: $I=0.05$, $0.1$, $0.2$ and $0.3$. This random perturbation has been done $n=5$ times to produce the following results, and to illustrate the reproductability of our approach. We first selected a scale radius corresponding to $4\%$ of the maximum bounding box size, and performed the global processing on all the meshes. From $I=0.05$ to $I=0.2$, the number of segmented texels and the result of the classification remains similar to the ground truth for all the runs. 

\begin{figure}
\includegraphics[width=0.9\linewidth]{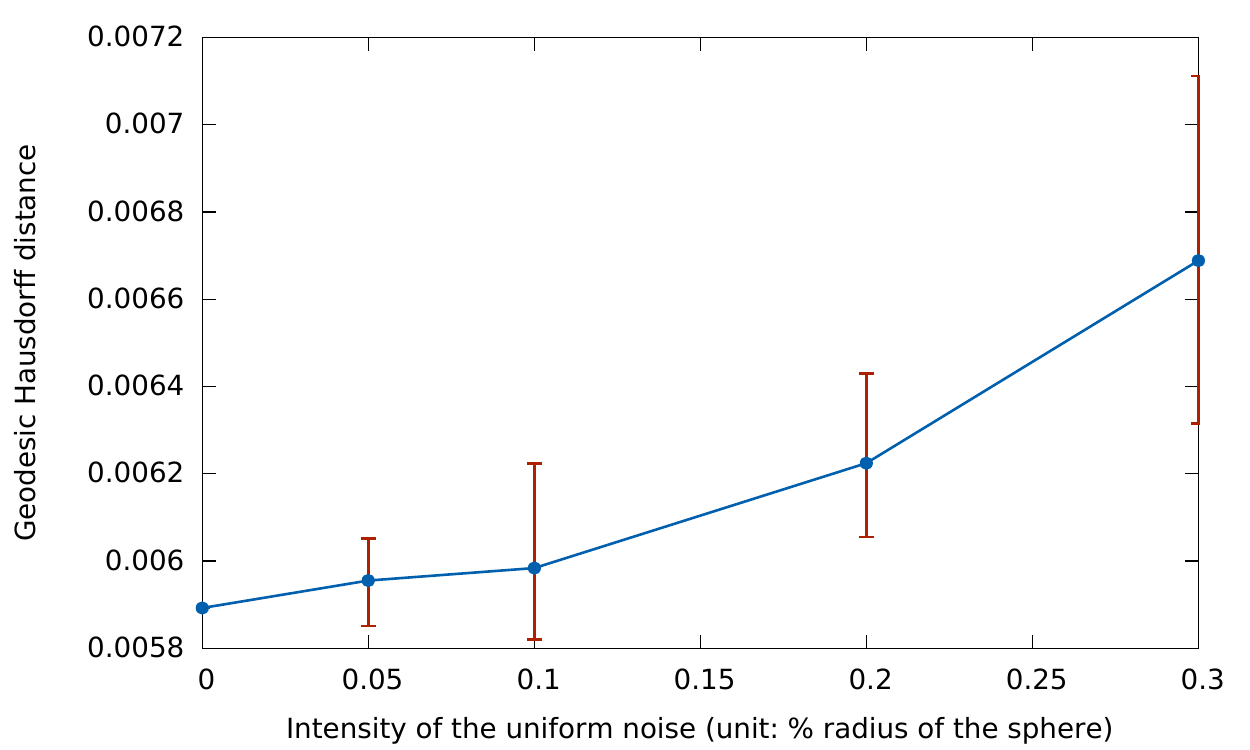}

\caption{Segmentation error in dependence of the noise level. Vertical red segments corresponds to the amplitude over the 5 runs at each intensity.}\label{fig:result-noise-2}
\end{figure}

However, when $I=0.3$ the segmentation generates some supplementary regions $E_n$ corresponding to highly noised background regions. By increasing the user scale to $6\%$ we were able to reduce to zero these supplementary regions. Figure~\ref{fig:result-noise-2} illustrates the stability of the segmentation using the Hausdorff distance introduced in section~\ref{sec:hausdorff}. The red intervals corresponds to the variation we measured for the 5 runs, while the blue dots corresponds to the mean over these 5 runs. The Hausdorff distance has been applied to compare the ground truth segmentation (see beginning of section~\ref{sec:evaluation}) with the generated segmentation. In case of $I=0.3$ where the noise can produce supplementary regions (depending on the chosen scale), we only considered the regions corresponding to the original ones.

\subsubsection{Influence of the mesh density}

\begin{figure}
 \includegraphics[width=.9\linewidth]{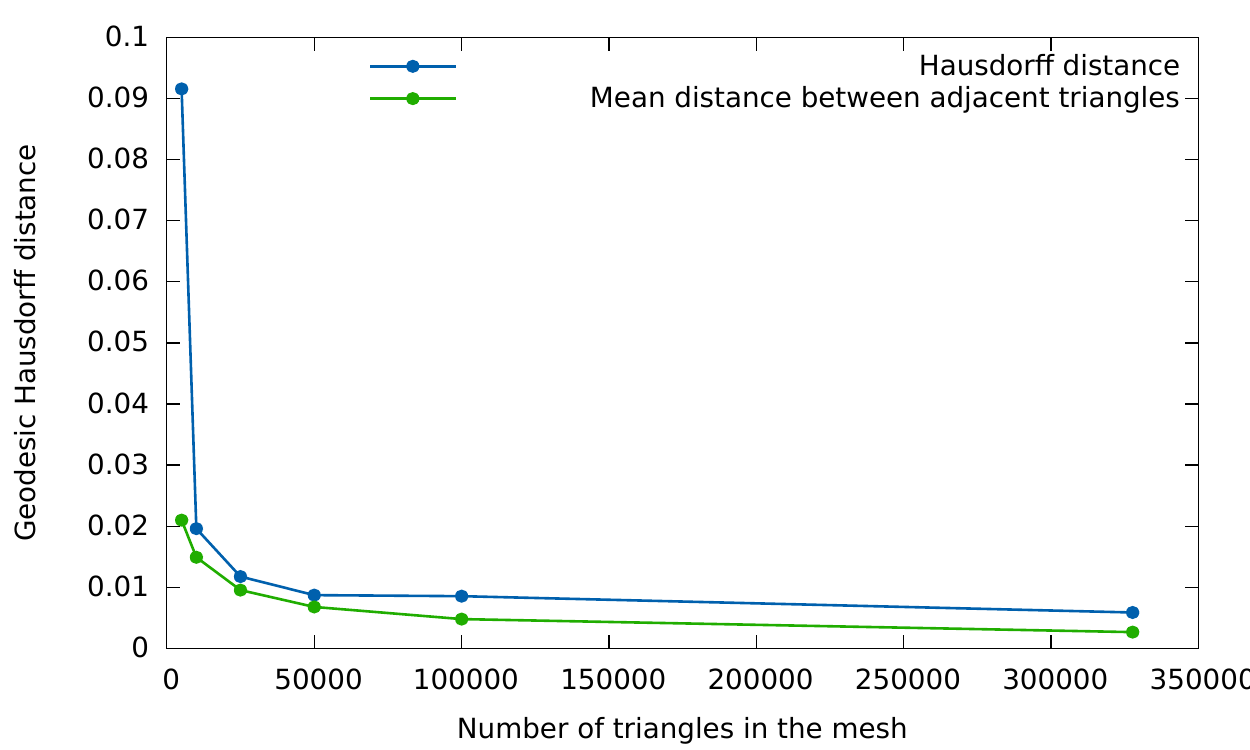}
 \caption{Segmentation error in dependence of the resolution.\label{fig:result-lowres}}
\end{figure}

Starting from the original mesh (Figure~\ref{fig:sphere-texture1}) with \np{327680} triangles, we simplified it at 5 different resolutions: \np{100000} triangles, \np{50000} triangles,  \np{25000} triangles, \np{10000} and \np{5000} triangles. The greyscale value of each vertex in simplified meshes has been obtained by considering the value associated to the closest vertex in the original mesh. Figure~\ref{fig:result-lowres} presents the segmentation error in dependence of the resolution.  To compare this error with the size of the triangles in each mesh, we plot a green line corresponding to the mean distance $\delta_{t}$ between adjacent triangles when the mesh contains $t$ triangles. 

Except for the last mesh with \np{5000} triangles, where geometric details are strongly distorted, the difference between the two lines is almost constant, and very small.
The error estimated by the Hausdorff distance is thus related to the resolution, almost corresponding to $\delta_{t}+ \delta_{327680}$. One possible interpretation of this error is the fact that our regions are defined by triangles, rather than the ground truth segmentation and the simplification has been done on the vertices.

\subsection{Results of the segmentation and classification}
In the previous section, we evaluated the presented approach on synthetic meshes. The results are satisfactory, but remains to be verified on meshes with more complex textures. We present here the results of the extraction and the classification of 3D texels on three popular textured meshes from the Shape Repository developed in the AIM@SHAPE project. Each segmentation illustration is presented jointly with the scale on the 3D mesh, the user scale is the radius of a blue disc.

\subsubsection{Classification results on textured meshes}

\begin{figure} 
        \subfloat[Bimba]{\label{fig:bimba} 
        \includegraphics[width=.25\linewidth]{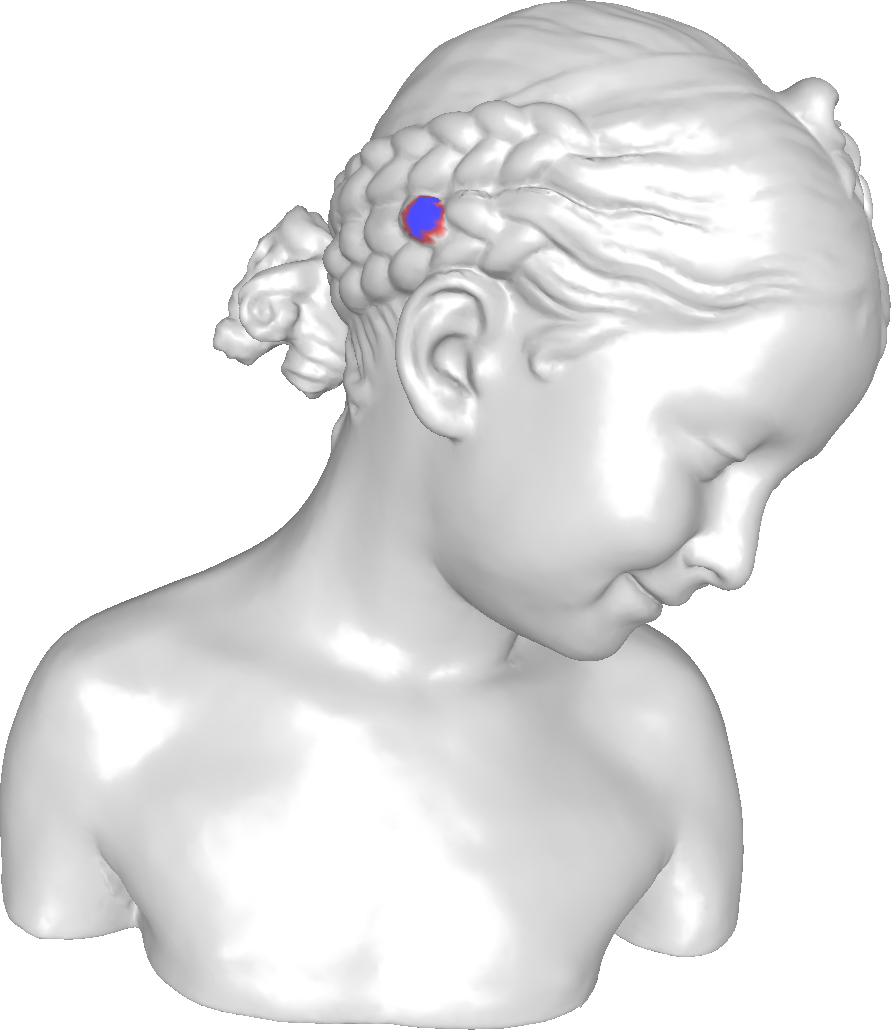} 
		\includegraphics[width=.25\linewidth]{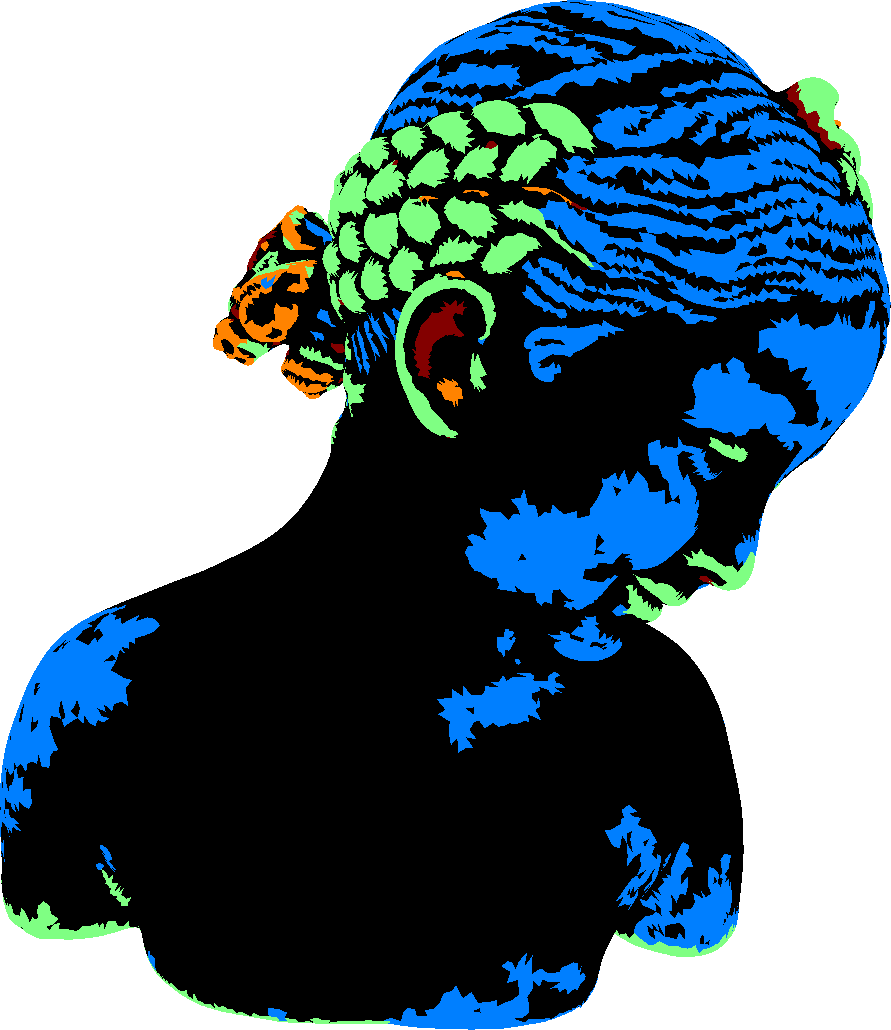}}  \hspace*{\stretch{1}}   	\textcolor{lightgray}{\vrule width 0.5mm}
        \subfloat[Dinosaur]{\label{fig:dinosaur} 
        \includegraphics[width=.2\linewidth]{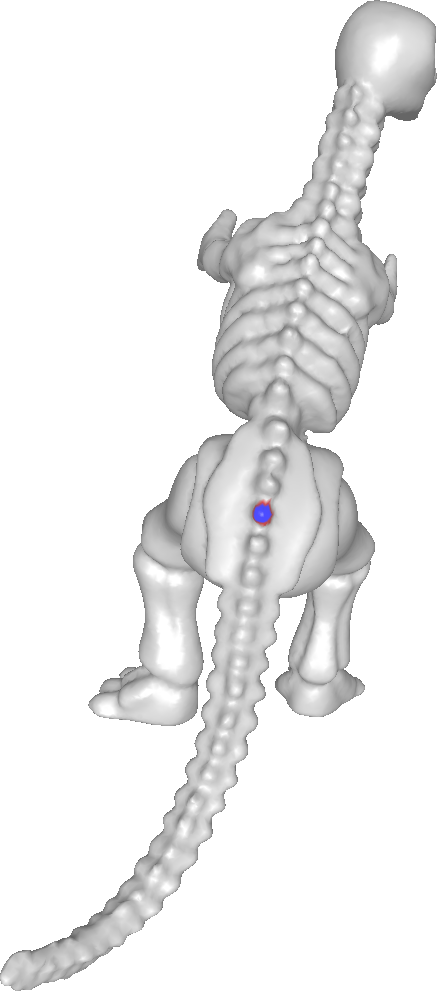} 
		\includegraphics[width=.2\linewidth]{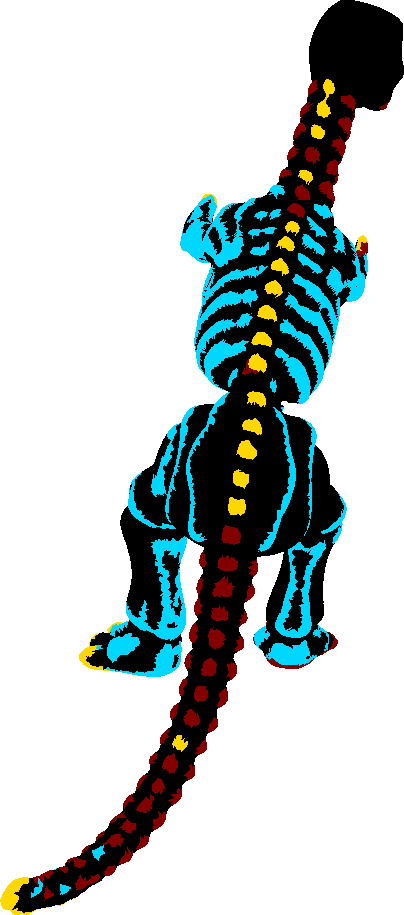}}
\textcolor{lightgray}{\hrule height 0.5mm}
    \vspace*{2mm}
    
     \subfloat[Lion]{\label{fig:tete_lion} 
        \includegraphics[width=.23\linewidth]{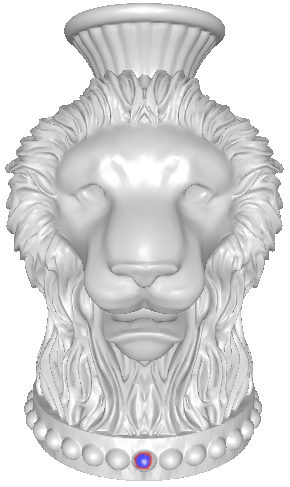} 
        \hspace*{\stretch{2}}
		\includegraphics[width=.23\linewidth]{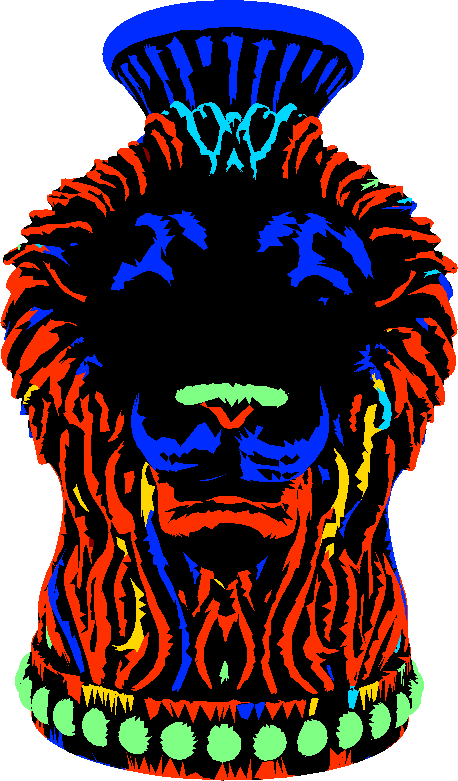}
		\hspace*{\stretch{2}}
        \includegraphics[height=.38\linewidth]{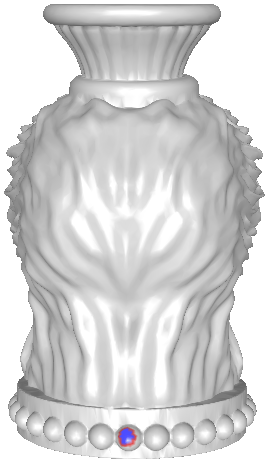}    
        \hspace*{\stretch{2}}
        \includegraphics[height=.38\linewidth]{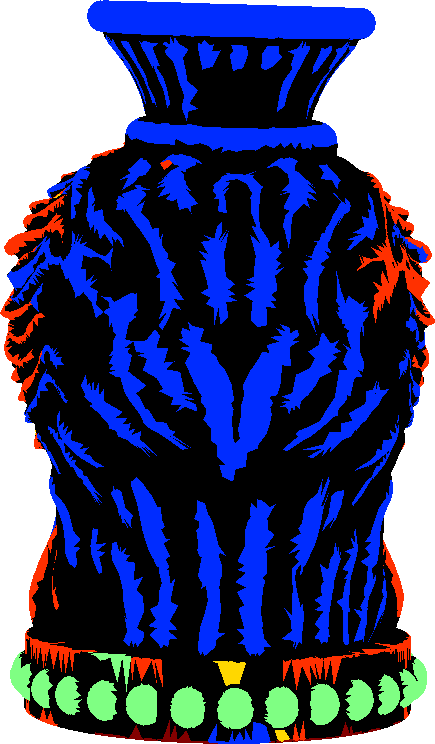} }

    \caption{Results of the classification on textured meshes. One color per class.}\label{fig:textured-meshes}
\end{figure}

We selected for the first mesh (Bimba, Figure~\ref{fig:bimba}) a fixed scale radius corresponding to the size of the braid node. The results of the segmentation are very interesting and meaningful: the braid nodes are gathered together, the bun nodes are gathered together and the rest of all waved lines of the hair together. The second mesh corresponds to the dinosaur (Figure~\ref{fig:dinosaur}): the vertebrae and the ribs are extracted correctly but the vertebrae are divided into two groups mainly because some of them are flatter than the other. The third mesh corresponds to the vase of lion head (Figure~\ref{fig:tete_lion}). The results are compelling and our approach succeed to separate the mane (red, green, sky blue), the rest of the fur (dark blue) and the balls of the base of the vase (green). In fact, the strips of the ornament above the vase are classified as the fur, the explication that we can give to that is that both of them correspond to stripes with different degree of undulation and length. 

It should also be noted that there is an over-segmentation and some regions that are segmented did not correspond to geometric textures but they are segmented because they correspond to 3D salient regions.

\subsubsection{Segmentation results on scale variation}

\begin{figure}
 \centering
 
 \hspace*{\stretch{1}}
 \includegraphics[width=.25\linewidth]{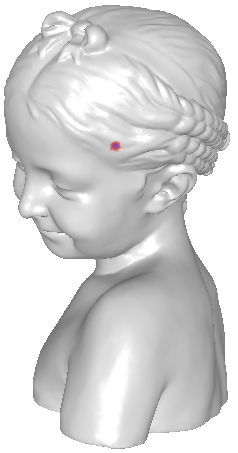}
 \hspace{\stretch{1}}
 \includegraphics[width=.25\linewidth]{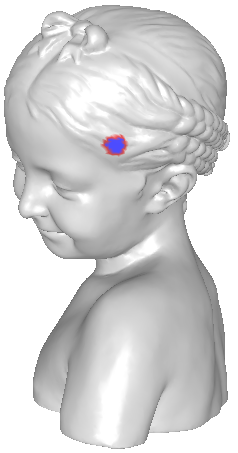}
 \hspace*{\stretch{1}}
 \includegraphics[width=.25\linewidth]{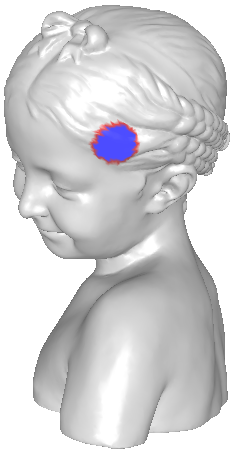}
 \hspace*{\stretch{1}} 

 \hspace*{\stretch{1}}
 \includegraphics[width=.25\linewidth]{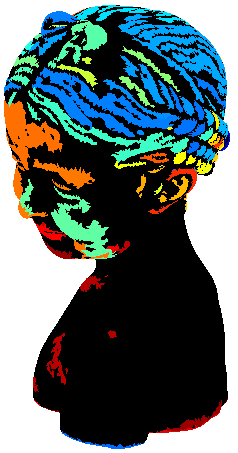}
 \hspace{\stretch{1}}
 \includegraphics[width=.25\linewidth]{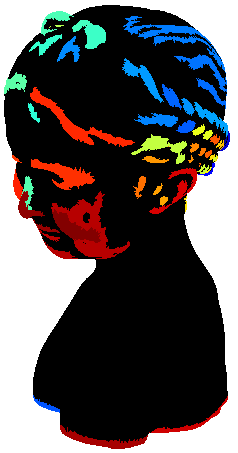}
 \hspace*{\stretch{1}}
 \includegraphics[width=.25\linewidth]{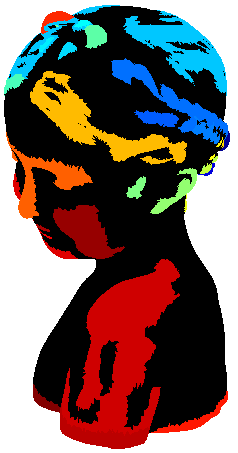}
 \hspace*{\stretch{1}}

 \caption{Segmentation result with three different radius on the Bimba mesh. The color of the regions is computed from the ID of the regions using a rainbow palette.\label{fig:multires}}
\end{figure}

\begin{figure}

\subfloat[A shape with smooth details.]{\label{fig:multires-smooth}
\begin{tabular}{r} 
 \includegraphics[height=.25\linewidth]{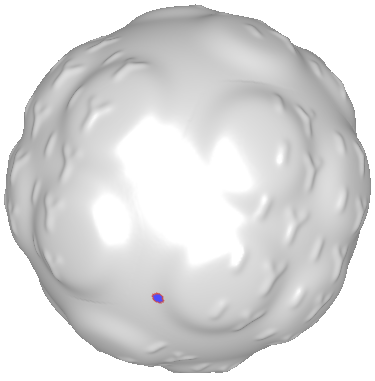} 
 \includegraphics[height=.25\linewidth]{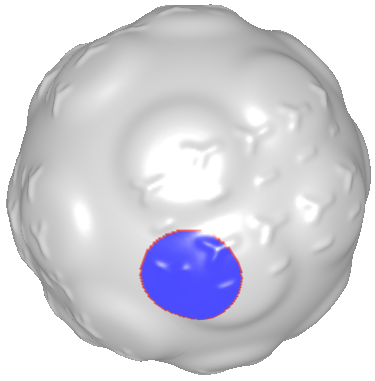} \\

 \includegraphics[height=.25\linewidth]{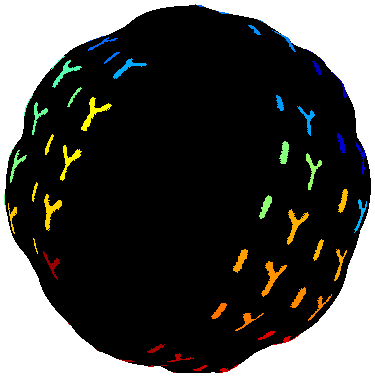}
 \includegraphics[height=.25\linewidth]{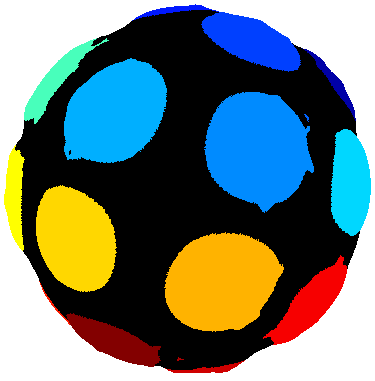}
 \end{tabular}}
\subfloat[A shape with sharp details.]{\label{fig:multires-sharp}
\begin{tabular}{l}  
\includegraphics[height=.25\linewidth]{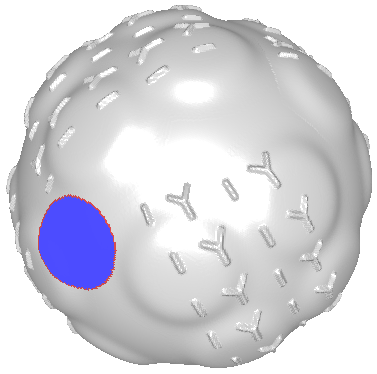} \\
\includegraphics[height=.25\linewidth]{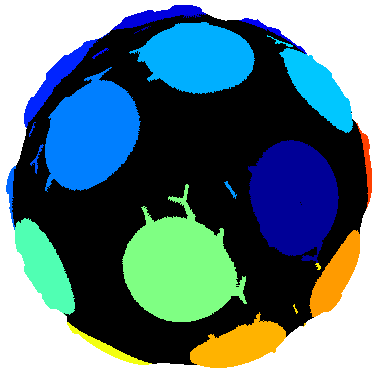}

\end{tabular}}

 \caption{Segmentation result with two different radius on two synthetic meshes.\label{fig:multires-a}}
\end{figure}

Figure~\ref{fig:multires} and \ref{fig:multires-a} illustrate the impact of the scale on the segmentation result, and the ability of our approach to select the desired level of detail. Figure~\ref{fig:multires} presents Bimba mesh with three different scales given by the user. The resulting segmentation is directly related to this scale: the thickness of the extracted details is very precise with the small radius, while the big radius produces large regions corresponding to big details of the shape, such as the braid or strands.

Figure~\ref{fig:multires-smooth} presents a synthetic object with two superimposed levels of details: big spheres and small strokes. Each of the selected scales gives a segmentation that focus on a level of details. Notice that the small details are not selected with the large radius even if they remain outside of the big details.
Depending of the sharpness of the small details, they can possibly be still visible at a big scale, as visible in Figure~\ref{fig:multires-sharp}.

\subsection{Application: semantic annotation of the 3D texels}
The new amount of visual data available in fields like satellite and bio-clinical imaging definitely calls for new paradigms whereupon knowledge engineering and computer vision issues must cooperate for the end-user benefits. Semantic technologies should play a key role for those new issues within the visual computing paradigms, they can offer promising approaches to image retrieval as they can map the image features to concepts. Consequently, they can answer the final user requirement and guarantee a better understanding and a better cooperation between the computer vision researchers and the end-users like the pathologists in the case of medical applications, the archaelogists for cultural heritage applications, the foresters for remote sensing applications, etc., \\
In this paper, we introduce a general application of semantic annotation of the 3D texels that will be used in our future work in retrieving and annotating 3D models in the field of Cultural Heritage. We should mention that our work can target many others applications of 3D object retrieval and texture analysis, the reason why we put the implementation under open source license  to benefit as many computer graphics and image processing applications as possible. \\

In brief, semantic annotation of a 3D object consists on associating to the object or a subpart of it a semantic concept. It is one of the main stage for several applications, such as serious games or mesh retrieval~\cite{Attene:2009}. Such annotation usually deals with large aspects of the shape: large regions related to the geometry (sphere, plane, protrusion) or the topology (extremity, genus, global structure)~\cite{Dietenbeck:2015}. In this section, we present one application of the geometric texture analysis as an original way to enrich the existing shape annotations. 

We described in section~\ref{sec:features} a series of features to characterize the geometry of the 3D texels. Each class of geometric texture $\mathcal{T}$ in the final classification is thus described by the set of all these features $\left\{\left\{f_{i,k}\right\}_k| R_i \in \mathcal{T}\right\}$. The main idea of this annotation part consists on identifying for each class the more significant features. A significant feature has to be understood in this section as a feature with maximal standard deviation ($\sigma(\cdot)$) over all the set of $\mathcal{R}$ all regions of the object, but with a minimal standard deviation over the regions of the current class. We modelized this property by the following measure, for a given class $\mathcal{T}$ and a given feature $i$:

\begin{equation}
S_{i, \mathcal{T}}= \frac{\sigma(\left\{f_{i,k} | R_k \in \mathcal{T}\right\})}{\sigma(\left\{f_{i,k} | R_k \in \mathcal{R}\right\})}. 
\end{equation}

Additionally to this significance evaluation, for each feature the mean value($\mu(\cdot)$) is computed on each region $V_{i, \mathcal{T}}=\mu(\left\{f_{i,k} | R_k \in \mathcal{T}\right\})$ as an overall estimation of feature values for this region. The semantic annotation of each texture $\mathcal{T}$ consists then in selecting the $n=5$ more significant features for which the feature value is semantically comprehensible. \\
To achieve this goal, we selected and designed the features introduced in section~\ref{sec:features} such that they are directly meaningful: the normalized area gives an intuition of the size, the local diameter has to be associated to the thickness of the shape, while the gaussian curvature mean gives an idea of the convexity of the regions. Using our expertise in geometry processing and a large number of evaluations on various meshes, we adjusted for each feature a series of intervals with semantic. A subpart of our semantic dictionary is shown in Table~\ref{fig:semantic-extract}, and the complete version is provided with our source code. We did not described all the intervals with annotations (e.g.\ squared perimeter to area ratio between 8 and 16) when the meaning appeared not explicit from our point of view.

\begin{table}
 \begin{tabular}{p{1.8cm}ccl}
 \textbf{Feature} & \textbf{left} & \textbf{right} & \textbf{semantic} \\
 \hline
 \multirow{3}{*}{\parbox{1.8cm}{Contour sphericity}} & 0 & \np{0.05} & circular shape\\
 & \np{0.05} & \np{0.15} & close to circular shape\\
 & \np{0.15} & \np{1} & non circular shape\\
 \hline
 \multirow{5}{*}{\parbox{1.8cm}{Gaussian curvature mean}} & $-\infty$&\np{-200} &very concave region\\
 & \np{-200} & \np{-50} &concave region \\
 &\np{-50} & \np{50} & flat region \\
 & \np{50} & \np{200} & convex region \\
 & \np{200} & $+\infty$ &very convex region \\
  \hline
  \multirow{4}{*}{\parbox{1.8cm}{Squared perimeter to area ratio}} & \np{0} & \np{4} & very bumped region \\
  & \np{4} & \np{8} & bumped region \\
  & \np{16} & \np{45} & non compact shape \\
  & \np{45} & $+\infty$ & very non compact shape \\
 \end{tabular}

 \caption{Part of a semantic dictionary with correspondence between feature values and annotations.}\label{fig:semantic-extract}
\end{table}

\label{sec:annotation}
\begin{figure}
 \raisebox{-0.5\height}{\includegraphics[width=.28\linewidth]{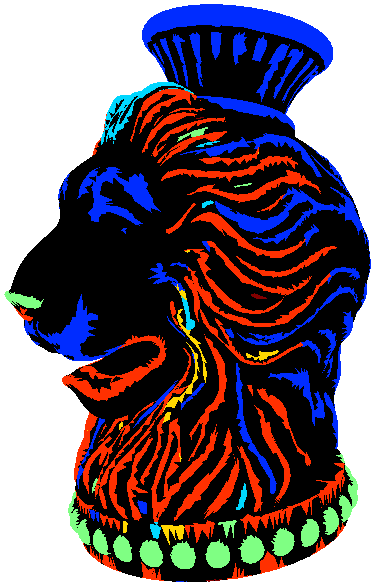}}
  \resizebox{0.72\linewidth}{!}{\begin{tabular}{clll}
  \textbf{Color} & \textbf{Feature} & \textbf{Value} & \textbf{Semantic} \\
%   \hline
%   \fcolorbox{black}{black}{\rule{0pt}{6pt}\rule{6pt}{0pt}} & background & & \\
  \hline
  \mycaption{5}{legend1-a} & Gaussian curvature mean & 11.3573& flat region \\
  &Gaussian curvature std deviation &172.210 & uniform concavity\\
  &PCA variance (3rd e.v.) &0.00800 & flat region\\
  & Shape index mean &0.48146 & saddle rut\\
  & Depth standard deviation & 0.04800 & flat region \\
  \hline
  \mycaption{5}{legend1-b} & Local diameter mean & 0.02714& very thin shape\\
  & Local diameter std deviation &0.01076 & regular local diameter\\
  & Area& 0.01853 & very small area\\
  & Contour sphericity & 0.19561 &non circular shape \\
  & Bounding sphere radius& 0.13183& small region \\
  \hline
  \mycaption{5}{legend1-c} &Perimeter length & 0.05808& Very small perimeter\\
  & Area& 0.02889 & very small area\\
  & Bounding sphere radius& 0.10355& small region\\
  & perimeter$^2$ to area ratio&17.8809 & non compact shape \\
  & Local diameter std deviation& 0.008663 & uniform local diameter \\
  \hline
  \mycaption{5}{legend1-d} & Local diameter mean & 0.00818& very thin shape \\
  & Local diameter std deviation& 0.00272 & uniform local diameter\\
  & Area& 0.00578 & very small area\\
  & Gaussian curvature std deviation& 213.984 & uniform concavity \\
  & Gaussian curvature mean& 308.566 & very convex region \\
    
 \end{tabular}}
 
 \caption{Semantic annotation of the lion (major classes).}\label{fig:annotation-lion}
\end{figure}

\begin{figure}
 \raisebox{-0.5\height}{\includegraphics[width=.28\linewidth]{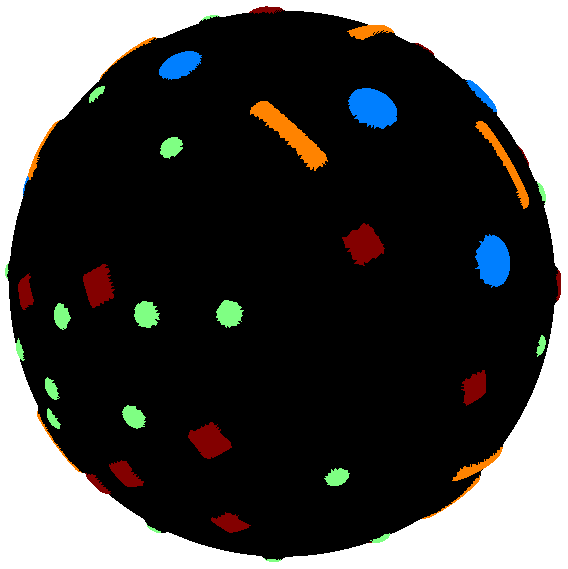}}
 \resizebox{0.72\linewidth}{!}{\begin{tabular}{clll}
  \textbf{Color} & \textbf{Feature} & \textbf{Value} & \textbf{Semantic} \\
%   \hline
%   \fcolorbox{black}{black}{\rule{0pt}{6pt}\rule{6pt}{0pt}} & background & & \\
  \hline
  \mycaption{5}{legend2-a} & Local diameter std deviation & 0.00800&uniform local diameter \\
  & PCA variance (2nd e.v.) & 0.19277&  two dominant axes \\
  & Contour sphericity & 0.01951 & circular shape \\
  & PCA variance (3rd e.v.) & 0.00501& flat region\\
  & Curvature index mean & 3.21860& flat region \\
  \hline
  \mycaption{5}{legend2-b} & PCA variance (1st e.v.) & 0.00571&uniform local diameter \\
  & Contour sphericity & 0.02657&  circular shape \\
  & PCA variance (2nd e.v.) & 0.47366 & two dominant axes \\
  & Bounding sphere radius & 0.02790& small region\\
  & Perimeter length & 0.03186& very small perimeter \\
  \hline
  \mycaption{5}{legend2-c} & PCA variance (1st e.v.) & 0.95250&one dominant axis \\
  & PCA variance (3rd e.v.) & 0.00393&  flat region \\
  & Bounding sphere radius & 0.27220 & non circular shape\\
  & perimeter$^2$ to area ratio & 24.8635& non compact shape\\
  & Curvature index mean & 2.67410& flat region \\
  \hline
  \mycaption{5}{legend2-d} & PCA variance (2nd e.v.) & 0.48590&two dominant axes \\
  & Contour sphericity & 0.07567&  close to circular shape \\
  & Local diameter std deviation & 0.02744 & regular local diameter\\
  & PCA variance (3rd e.v.) & 0.00948& flat region\\
  & Perimeter length & 0.27550& small perimeter \\
 \end{tabular}}
 \caption{Semantic annotation of a synthetic shape}\label{fig:annotation-synthetic}
\end{figure}

The semantic annotation of the four major classes resulting from the mesh segmentation of the lion vase are presented in Figure~\ref{fig:annotation-lion}. As we can see, blue regions correspond to the flat strips, they were detected and annotated. The balls of the base of the vase are annotated as small regions, with non-compact shape, small area and small perimeter. The strips of the fur are well annotated as thin shape with very small area, uniform concavity and very convex region. 
Figure~\ref{fig:annotation-synthetic} gives the result of the annotation on a synthetic shape, with hand made geometric details. The four classes are correctly annotated and detected: the rectangles (orange regions) are labeled as ``one dominant axis'', while the circles (both small and big) are labeled as ``circular shape'' and the squares as ``close to circular shape''. 

First annotation results are promizing. They should be combined with already existing approaches to produce high level semantic descriptions, using for example a modelization of these annotations
with an ontology enriched with inference rules. A future work should be to consolidate the interval bounds of the semantic dictionary, for example with the help of an experimental protocol with user feedback.

\section{Conclusion and future work}
In this work, we present a new scale-aware geometric texture segmentation and classification tool. The proposed approach extracts 3D texels (or shape details) having the same geometrical aspect. We highlighted the efficiency of the method to extract geometric details from natural and synthetic meshes, with respect to several user-defined scales. Furthermore, we conducted experimental studies to investigate the efficiency of the method after noise addition and mesh simplification. Finally, we presented a practical application to the semantic annotation of 3D objects using the proposed method. We also published an open source implementation of the presented approach, developed using CGAL library, to facilitate the evaluation and extension of our work. \\

A first direction of our future work is to use our approach of 3D geometric salient textures analysis on retrieving and recognizing data from 3D scanning of cultural heritage. The semantic knowledge will be formalized in tree-structure based ontology as described in ~\cite{Dietenbeck:2015,Dietenbeck2:2015,Othmani:2010} for a better organization of the corresponding information and for reasoning. A second interesting direction for our future work is to extend the method to colored meshes and study the joint effect of geometrical and image-based textures on 3D meshes analysis. 
An original aspect of our work is the semantic annotation of meshes using geometric textures. We plan to integrate this semantic annotation as an elementary ingredient of more complex segmentation frameworks, using for example ontologies to describe these elementary annotations~\cite{Dietenbeck:2015}. 
One other straightforward extension of this work will be to produce a complete segmentation of the mesh, removing background label for the final segmentations to produce a partitionning of the mesh in areas with similar local geometric details. Vorono\"i diagrams on the surface of the mesh using 3D texels as sites is trivial approach we will have to explore and improve.
We also plan to work on future applications of the proposed method for geometric details transfer in mesh editing applications.
%\begin{acknowledgements}
%\end{acknowledgements}


\begin{thebibliography}{}
\bibitem{Uhl:2015}
Uhl, A. and Wimmer, G.: A systematic evaluation of the scale invariance of texture recognition methods. In Pattern Analysis and Applictions, 18: 945 (2015). 

\bibitem{Lavoue:2009}
Lavou\'e , Guillaume: Local Roughness Measure for 3D Meshes and its Application to Visual Masking. In ACM Transactions on Applied Perception, Vol. 5, No. 4, Article 21, (2009).

\bibitem{Otsu:1979}
Otsu, Nobuyuki: A threshold selection method from gray-level histograms. In IEEE Transactions on Systems, Man and Cybernetics, Vol. 9, pp 62--66 (1979).

\bibitem{Koenderink:1992}
Koenderink, Jan J and van Doorn, Andrea J: Surface shape and curvature scales. In Image and vision computing, Vol. 10, pp 557--564 (1992).

\bibitem{Osada:2002}
Osada, Robert and Funkhouser, Thomas and Chazelle, Bernard and Dobkin, David: Shape distributions.In ACM Transactions on Graphics, Vol. 21, pp 807--832 (2002).

\bibitem{Mortara:2004}
Mortara, Michela and Patan{\'e}, Giuseppe and Spagnuolo, Michela and Falcidieno, Bianca and Rossignac, Jarek: Blowing bubbles for multi-scale analysis and decomposition of triangle meshes. In Algorithmica, Vol. 38, pp 227--248 (2004).

\bibitem{Lee:2005}
Lee, Chang Ha and Varshney, Amitabh and Jacobs, David W: Mesh saliency. In ACM transactions on graphics, Vol. 24, pp 659--666 (2005).

\bibitem{Zhou:2006}
Zhou, Kun and Huang, Xin and Wang, Xi and Tong, Yiying and Desbrun, Mathieu and Guo, Baining and Shum, Heung-yeung: Mesh Quilting for Geometric Texture Synthesis. In ACM Transactions on Graphics, Vol. 25, pp 690--697 (2006).

\bibitem{Gal:2006}
Gal, Ran and Cohen-Or, Daniel: Salient geometric features for partial shape matching and similarity. In ACM Transactions on Graphics, Vol. 25, pp 130--150 (2006).

\bibitem{Toledo:2008}
De Toledo, Rodrigo and Wang, Bin and Levy, Bruno: Geometry Textures and Applications. In Computer Graphics Forum, Vol. 27, pp 2053--2065 (2005).

\bibitem{Shapira:2008}
Shapira, Lior and Shamir, Ariel and Cohen-Or, Daniel: Consistent mesh partitioning and skeletonisation using the shape diameter function. In The Visual Computer, Vol. 24, pp 249--259 (2008).

\bibitem{Vallet:2008}
Vallet, Bruno and Levy, Bruno: Spectral Geometry Processing with Manifold Harmonics. In Computer Graphics Forum, Vol. 27, pp 251--260 (2008).

\bibitem{Andersen:2009}
Andersen, Vedrana and Desbrun, Mathieu and Bærentzen, J. Andreas and Henrik, Aanæs: Height and Tilt Geometric Texture. In Internation Symposium on Visual Computing, Lecture Notes in Computer Science, Vol. 5875, pp 656--667 (2009).

\bibitem{Attene:2009}
Attene, Marco and Robbiano, Francesco and Spagnuolo, Michela and Falcidieno, Bianca: Characterization of 3D shape parts for semantic annotation. In Computer-Aided Design, Vol. 41, pp 756–-763 (2009).

\bibitem{Alliez:2009}
Alliez, Pierre and Tayeb, Stephane and Wormser, Camille: AABB Tree. In CGAL 3.5 edition (2009).

\bibitem{Alhashim:2012}
Alhashim, Ibraheem and Zhang, Hao and Liu, Ligang: Detail-replicating shape stretching. In The Visual Computer, Vol. 28, pp 1--14 (2012).

\bibitem{Othmani:2013}
Othmani, Ahlem and Voon, Lew FC Lew Yan and Stolz, Christophe and Piboule, Alexandre: Single tree species classification from Terrestrial Laser Scanning data for forest inventory. In Pattern Recognition Letters, Vol. 34, pp 2144--2150 (2013).

\bibitem{Guy:2014}
Guy, Emilie and Thiery, Jean-Marc and Boubekeur, Tamy: SimSelect: Similarity-based selection for 3D surfaces. In Computer Graphics Forum, Vol. 33, pp 165--173 (2014).

\bibitem{Othmani:2014}
Othmani, Ahlem: Identification automatis\'ee des esp\`eces d`arbres dans des scans lasers 3\textsc{D} r\'ealis\'es en for\^et. In PhD thesis, (2014).

\bibitem{Song:2014}
Song, Ran and Liu, Yonghuai and Martin, Ralph R. and Rosin, Paul L.: Mesh saliency via spectral processing. In ACM Transactions on Graphics, Vol. 33, pp 6:1--6:17 (2014).

\bibitem{Taubin:2000}
Taubin, Gabriel: Geometric Signal Processing on Polygonal Meshes. In Eurographics State of the Art Report, (2000).

\bibitem{LihiZelnikManorPPerona:2004}
Zelnik-Manor, Lihi and Perona, Pietro: Self-tuning spectral clustering. In Advances in neural information processing systems, pp 1601--1608 (2004).

\bibitem{Lai:2005}
Lai, Yu-Kun and Hua, Shi-Min and Gu, Xianfeng and Martin, Ralph. R.: Geometric Texture Synthesis and Transfer via Geometry Images. In Proc. ACM Symp. on Solid and Physical Modeling, pp 15--26 (2005).

\bibitem{Cheng:2007}
Cheng, Zhi-Quan and Dang, Gang and Jin, Shi-Yao: A meaningful mesh segmentation based on local self-similarity analysis. In IEEE International Conference on Computer-Aided Design and Computer Graphics, pp 288--293 (2007).

\bibitem{Zaharescu:2009}
Zaharescu, Andrei and Boyer, Edmond and Varanasi, Kiran and Horaud, Radu: Surface feature detection and description with applications to mesh matching.In IEEE Conference on Computer Vision and Pattern Recognition, pp 373--380 (2009).

\bibitem{Tabia:2014}
Tabia, Hedi and Laga, Hamid and Picard, David and Gosselin, Philippe-Henri: Covariance descriptors for 3D shape matching and retrieval. In IEEE Conference on Computer Vision and Pattern Recognition, pp 4185--4192 (2014). 

\bibitem{Dietenbeck:2015}
Dietenbeck, T. and Othmani, A. and Attene, M. and Favreau J.-M.: A Framework for Mesh Segmentation and Annotation using Ontologies. In Extraction et gestion des connaissances, pp 275--286 (EGC'2015).

\bibitem{Dietenbeck2:2015}
Dietenbeck, T. and Torkhani, F. and Othmani, A. and Attene, M. and Favreau, J. - M. : Multi-layer ontologies for integrated 3D shape segmentation and annotation. In Advances in Knowledge Discovery and Management (AKDM-6), 2016.

\bibitem{Othmani:2010}
Othmani, A. and Meziat, C. and Loménie, N. : Ontology-driven image analysis for histopathological images. In International Symposium on Visual Computing, pp 1--12 (2010).

\end{thebibliography}


\end{document}